\documentclass[sigconf,nonacm]{acmart}
\AtBeginDocument{%
  }

\setcopyright{none}
\acmDOI{}
\acmISBN{}

\usepackage{tabularx}
\usepackage{graphicx}
\usepackage{xcolor}
\usepackage{pifont}
\usepackage{tikz} 
\usepackage{enumitem}
\usepackage{multirow}
\usepackage{caption}
\usepackage{subcaption}
\hypersetup{pdfauthor={Xuan Wang, Jianan Jiang, Rongqian Chen, Tian Lan, Bin Li}}
\usetikzlibrary{shapes.geometric, arrows.meta, positioning, fit, backgrounds}

\begin{document}
\raggedbottom

\title{Too Hard, Too Easy, Just Right: Performance-Driven Challenge Adaptation in Virtual Reality Firefighter Training}


\author{Xuan Wang}
\authornote{Both authors contributed equally to this research.}
\affiliation{%
  \institution{The Pennsylvania State University}
  \city{State College}
  \state{PA}
  \country{USA}
}
\email{xuanwang@psu.edu}

\author{Jianan Jiang}
\authornotemark[1]
\affiliation{%
  \institution{The Pennsylvania State University}
  \city{State College}
  \state{PA}
  \country{USA}
}
\email{jnjiang@psu.edu}

\author{Rongqian Chen}
\affiliation{%
  \institution{The George Washington University}
  \city{Washington}
  \state{DC}
  \country{USA}
}
\email{rongqianc@gwu.edu}

\author{Tian Lan}
\affiliation{%
  \institution{The George Washington University}
  \city{Washington}
  \state{DC}
  \country{USA}
}
\email{tlan@gwu.edu}

\author{Bin Li}
\affiliation{%
  \institution{The Pennsylvania State University}
  \city{State College}
  \state{PA}
  \country{USA}
}
\email{binli@psu.edu}

\renewcommand{\shortauthors}{Wang et al.}

\begin{abstract}

Effective firefighter training requires practice under dynamic and stressful emergency conditions, yet real-world live-fire exercises are costly, risky, and difficult to scale. Virtual Reality (VR) offers a safe and controllable alternative, but many existing VR firefighter training systems rely on fixed scenarios or provide limited difficulty adaptation, making it difficult to maintain an appropriate level of challenge for skill development as trainees' performance changes over time. To address this limitation, we present ProACT, a performance-driven adaptive training system that regulates scenario difficulty through closed-loop interactions between trainee performance and simulated fire dynamics. Specifically, ProACT combines a global module that continuously adjusts scene-level fire spread dynamics in real time based on trainee performance with a local module that periodically updates object-level flammability across the environment. Our evaluation shows that ProACT improves both objective training performance and subjective user experience compared with existing adaptive and non-adaptive baselines. We further discuss how adaptive immersive environments can extend beyond firefighting to broader training domains.

\end{abstract}

\ccsdesc[500]{Human-centered computing~Ubiquitous and mobile computing systems and tools}
\ccsdesc[500]{Human-centered computing~Human computer interaction (HCI)}
\ccsdesc[500]{Human-centered computing~Virtual reality}
\ccsdesc[500]{Computing methodologies~Artificial intelligence}

\keywords{Immersive Training, Firefighter Training, Dynamic Difficulty Adjustment, Large Language Model}


\begin{teaserfigure}
    \centering
    \includegraphics[width=\linewidth]{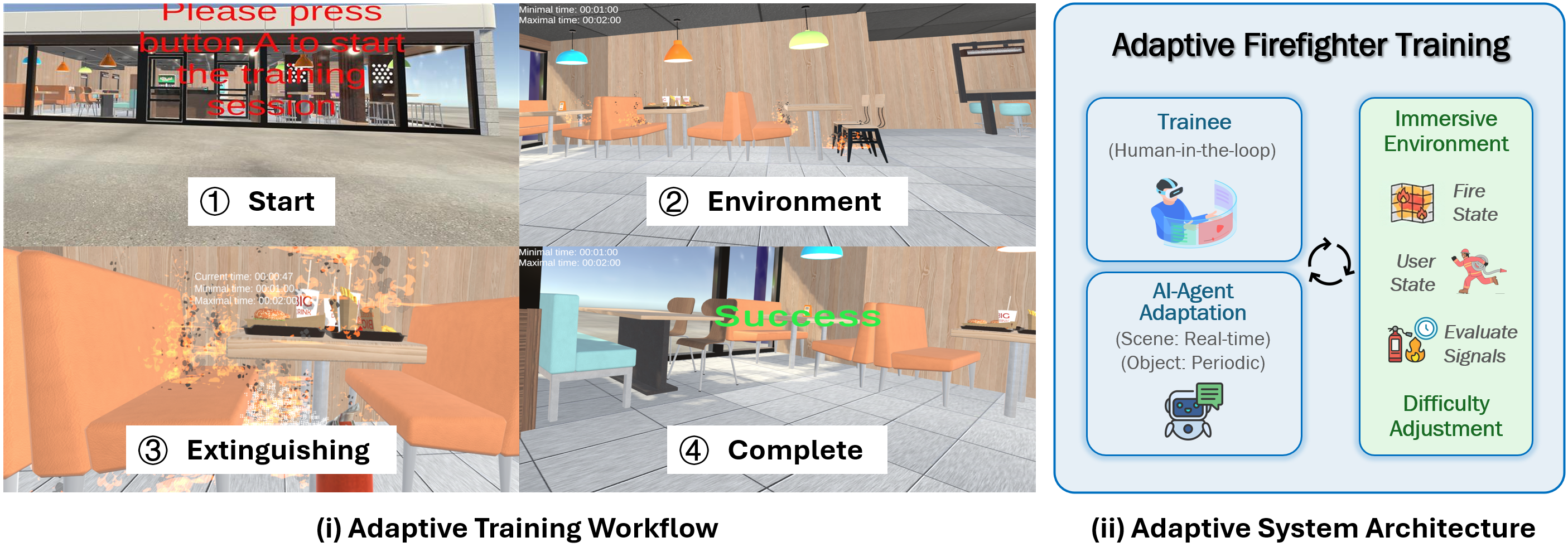}
    \Description{Four stages of the VR training workflow and the human-in-the-loop architecture linking the trainee, the virtual environment, and the global and local adaptation modules.}
    \caption{(i) VR training workflow consists of four stages: \ding{172} \textbf{Start}, where trainees initiate the session; \ding{173} \textbf{Environment}, where trainees explore the scene and identify potential hazards; \ding{174} \textbf{Extinguishing}, during which users actively suppress the spreading fire; and \ding{175} \textbf{Complete}, which indicates successful task completion. (ii) System architecture illustrating how the trainee (human-in-the-loop) interacts with the immersive environment, while the AI-agent performs scene-level real-time regulation of fire spread dynamics and object-level periodic updates of flammability to maintain an appropriate challenge level during training.}
    \label{fig:teaser}
\end{teaserfigure}


\maketitle

\section{Introduction}

Virtual Reality (VR) has emerged as a practical platform for immersive training in high-risk domains by enabling safe, repeatable, and controllable simulation of hazardous scenarios~\cite{scorgie2024virtual,killeen2024stand,chen2025bridging}. Firefighter training is a particularly important application of VR~\cite{wheeler2021human,engelbrecht2019swot,mehta2025effectiveness}, because preparing firefighters for real incidents requires practice under dynamic and stressful conditions, such as extreme heat, dense smoke, limited visibility, and rapidly changing fire behavior. However, conducting such training in the real world is inherently dangerous, resource-intensive, and difficult to scale~\cite{nfpa_injury_report,bls_firefighters}. In practice, ensuring safety during live-fire training imposes substantial operational constraints, making such exercises costly, logistically demanding, and limited in frequency. \textit{VR therefore offers an opportunity to provide trainees with more accessible exposure to hazardous firefighting scenarios while preserving experimental control and user safety.}

Prior research on VR firefighter training has demonstrated the potential of immersive simulation for tasks such as search-and-rescue, evacuation, and incident command~\cite{vaughan2016overview,oliveira2024exploring}. However, many existing systems still rely on static, pre-defined scenarios, where difficulty is fixed before training or manually adjusted by instructors~\cite{backlund2007sidh,cha2012virtual}. Such approaches limit the ability of VR training systems to respond to trainees whose performance changes during a session. From a learning perspective, effective skill acquisition depends on aligning task difficulty with individual proficiency~\cite{guadagnoli2004challenge,bonaiuto2016optimal}: scenarios that are too easy may fail to promote improvement, while scenarios that are too difficult may overwhelm trainees and prevent meaningful practice. This issue is especially important in firefighter training, where challenge is not simply a property of a pre-authored scenario. Instead, it emerges from the interaction among fire propagation, object ignition, spatial layout, visibility, and trainee actions. \textit{Therefore, adaptive VR firefighter training should adjust simulated fire dynamics in response to trainees' evolving performance.}

While recent work has explored adaptive mechanisms in VR and training systems~\cite{verniani2024features,zahabi2020adaptive}, performance-driven regulation of dynamic firefighting scenarios remains an emerging capability rather than a pervasive component of VR firefighter training. Designing such systems introduces three key challenges:

\begin{itemize}
    \item[\textbf{C1}] \textbf{Real-time Challenge Regulation.} Training environments must translate evolving trainee performance into timely adjustments of scenario difficulty. This requires mapping task-level signals, such as elapsed time, extinguishing progress, and fire evolution, to corresponding changes in simulated fire behavior.
    \item[\textbf{C2}] \textbf{Precise Environment Regulation.} Scenario difficulty is shaped by multiple interacting factors, including overall fire progression, object-specific ignition behavior, spatial layout, and contextual flammability. Effective personalization requires fine-grained control of environment dynamics rather than coarse or uniform difficulty adjustments.
    \item[\textbf{C3}] \textbf{Stable yet Responsive Adaptation.} Adaptive systems must balance responsiveness to trainee performance with the stability of the simulated environment. Overly reactive adaptation may introduce abrupt or unrealistic changes in fire behavior, while delayed adaptation may fail to correct scenarios that become too easy or too difficult.
\end{itemize}

To address these challenges, we present \textbf{ProACT}, a \textbf{P}erformance-driven \textbf{A}daptive \textbf{C}hallenge \textbf{T}raining system that regulates scenario difficulty through real-time interactions between trainee performance and environment dynamics. ProACT monitors task-level signals such as elapsed time, extinguishing progress, and fire evolution, and adapts simulated fire behavior to maintain an appropriate challenge level. Specifically, our system operates through two complementary mechanisms. The \textit{Fire Dynamics Regulation Module} regulates global fire dynamics by continuously adjusting scene-level spread probability in real time according to trainee performance. The \textit{Semantic Scene Reasoning Module} supports local adaptation by using a large language model to infer object-level flammability from scene inputs and periodically update ignition likelihood across the VR environment. This design enables coordinated control across spatial and temporal scales, supporting both responsive and stable adaptation. By treating the trainee as a human-in-the-loop participant whose performance shapes the environment, ProACT enables online personalization without requiring explicit user modeling or manual retuning. In summary, this paper makes the following contributions:

\begin{itemize}
    \item We introduce \textbf{ProACT}, a performance-driven VR firefighter training system that treats simulated fire behavior as an adaptive environment variable and continuously regulates scenario difficulty based on real-time trainee performance.
    \item We propose a multi-scale challenge adaptation approach that combines real-time global fire spread regulation with periodic object-level flammability estimation, enabling coordinated control over both global fire dynamics and local object ignition behavior.
    \item Our evaluation demonstrates that performance-driven adaptive regulation provides measurable benefits for immersive training, including higher on-target completion rates, improved adherence to target challenge ranges, more appropriate perceived mental effort, and stronger perceptions of challenge appropriateness.
\end{itemize}

\section{Related Work}
\subsection{Virtual and Immersive Training Systems}
Immersive training systems have been widely explored as safe and controllable media for practicing skills that are difficult, expensive, or hazardous to rehearse in the real world, particularly in embodied and task-oriented settings~\cite{zahabi2020adaptive,butaslac2022systematic}. Prior work demonstrates this potential across domains, including open-ended machining practice beyond rigid tutorials~\cite{studer2024open}, in-situ basketball tactic training~\cite{cheng2024viscourt}, and emergency service training~\cite{pietschmann2024use}. In emergency-response settings, VR has also been used to simulate realistic stressors and operational contexts for firefighter training~\cite{engelbrecht2019swot,mehta2025effectiveness,clifford2019creating}. These studies suggest that immersive environments can support procedural rehearsal and provide controlled exposure to realistic emergency-response conditions. Beyond scenario simulation, immersive training systems increasingly function as instructional interfaces that guide users through feedforward cues, corrective feedback, and multimodal interaction. For example, prior work has explored visual corrective feedback for motor skill training in MR~\cite{diller2022visual}, broader design spaces for combining feedforward and feedback in XR-based motion guidance systems~\cite{yu2024design}, mentor-guided learning with visual and haptic feedback~\cite{lebrun2025mentor}, mixed reality rehabilitation interfaces that integrate speech, hand interaction, and embodied practice~\cite{wang2024mrehab}, and VR-based balance training with multi-sensory feedback~\cite{lin2025vrshuttlecock}. These efforts position immersive environments as interactive training systems rather than passive simulation displays. However, much of this work focuses on guidance, feedback, or interaction support, leaving less explored how the simulated hazard environment itself can be dynamically regulated to maintain appropriate training challenge.

\subsection{Dynamic and Adaptive VR Training}
Dynamic difficulty adjustment (DDA) has long been studied in games and interactive systems as a mechanism for maintaining an appropriate balance between task challenge and user capability. Foundational work emphasized that adaptive difficulty should respond to player performance in real time while remaining unobtrusive, fair, and consistent with user agency~\cite{zohaib2018dynamic,hunicke2005case}. Later research operationalized these principles through explicit difficulty modeling and player modeling~\cite{aponte2011measuring,missura2009player}, learned behavior models for long-term personalized balancing~\cite{pfau2020enemy,pfau2022dungeons}, procedural mission and level generation tailored to player skill~\cite{zook2012skill,biemer2023dynamic}, and physiological adaptation based on user state~\cite{moschovitis2022keep}. This literature provides a strong foundation for modeling challenge as a dynamic relationship between user capability, task demands, and system response. Motivated by similar goals, adaptive virtual reality training systems increasingly adopt closed-loop mechanisms that personalize training using performance signals and, in some cases, behavioral or physiological measures~\cite{zahabi2020adaptive,correia2024adaptive}. Prior systems have explored adaptation in feedback, interface behavior, interaction mappings, and training tasks to improve training outcomes and user experience~\cite{aguilar2023design,liao2025focus,chu2025design}. However, these approaches have largely focused on the instructional layer --- for example, feedback, interface features, interaction mappings, or task difficulty --- rather than the simulated environment itself. This limitation is particularly important in high-risk emergency-response training, where maintaining an appropriate challenge level may require the hazard environment to respond dynamically to trainee performance. In firefighter VR research, prior work has primarily focused on feasibility, ecological validity, stress induction, and human factors~\cite{engelbrecht2019swot,wheeler2021human,mehta2025effectiveness,clifford2019creating}, with limited exploration of performance-driven adaptive scenario control. Our work builds on both DDA and adaptive training literature by treating the hazard environment itself as the adaptive target, enabling real-time regulation of fire behavior in VR firefighter training.

\subsection{Generative AI and Context-Aware Intelligence}
Recent research shows that generative AI can support context-rich human-centered tasks by interpreting multimodal inputs, assisting decision-making, and generating situated responses. Prior work has used human-in-the-loop AI to assist expert sensemaking of multimodal data~\cite{li2025vital}, support creative workflows~\cite{jiang2024haigen}, enable reflection and planning around personal health data~\cite{chopra2025engagements}, and provide personalized AI-driven journaling experiences~\cite{nepal2024mindscape}. These studies suggest that generative AI can augment interpretation, planning, and decision-making in complex human-centered settings. This trend has increasingly extended to immersive systems. In VR environments, prior work has explored AI-assisted guidance and instruction, including mixed-reality learning assistants and generative systems for authoring context-aware AR instructions~\cite{chen2025origami,shi2025caring}. Other research has introduced LLM-based agents in virtual environments, enabling more flexible and engaging interactions than predefined scripted behaviors~\cite{wan2024building,jiang2026partinteractor}. Recent systems have also leveraged contextual signals such as gaze, dialogue history, and environmental context to improve interaction understanding and adapt mixed-reality interfaces to user needs~\cite{lee2024gazepointar,li2024situationadapt,cho2024minexr}. These works highlight the potential of combining generative AI with contextual intelligence to make immersive systems more adaptive, responsive, and personalized. However, existing AI-enhanced immersive systems rarely use generative models to regulate training environment dynamics. ProACT instead uses LLM-based semantic reasoning to estimate object-level flammability and adapt local ignition behavior.


\section{Adaptive Firefighter Training System}
\subsection{System Overview}

\begin{figure*}[t]
    \centering
    \includegraphics[width=.8\linewidth]{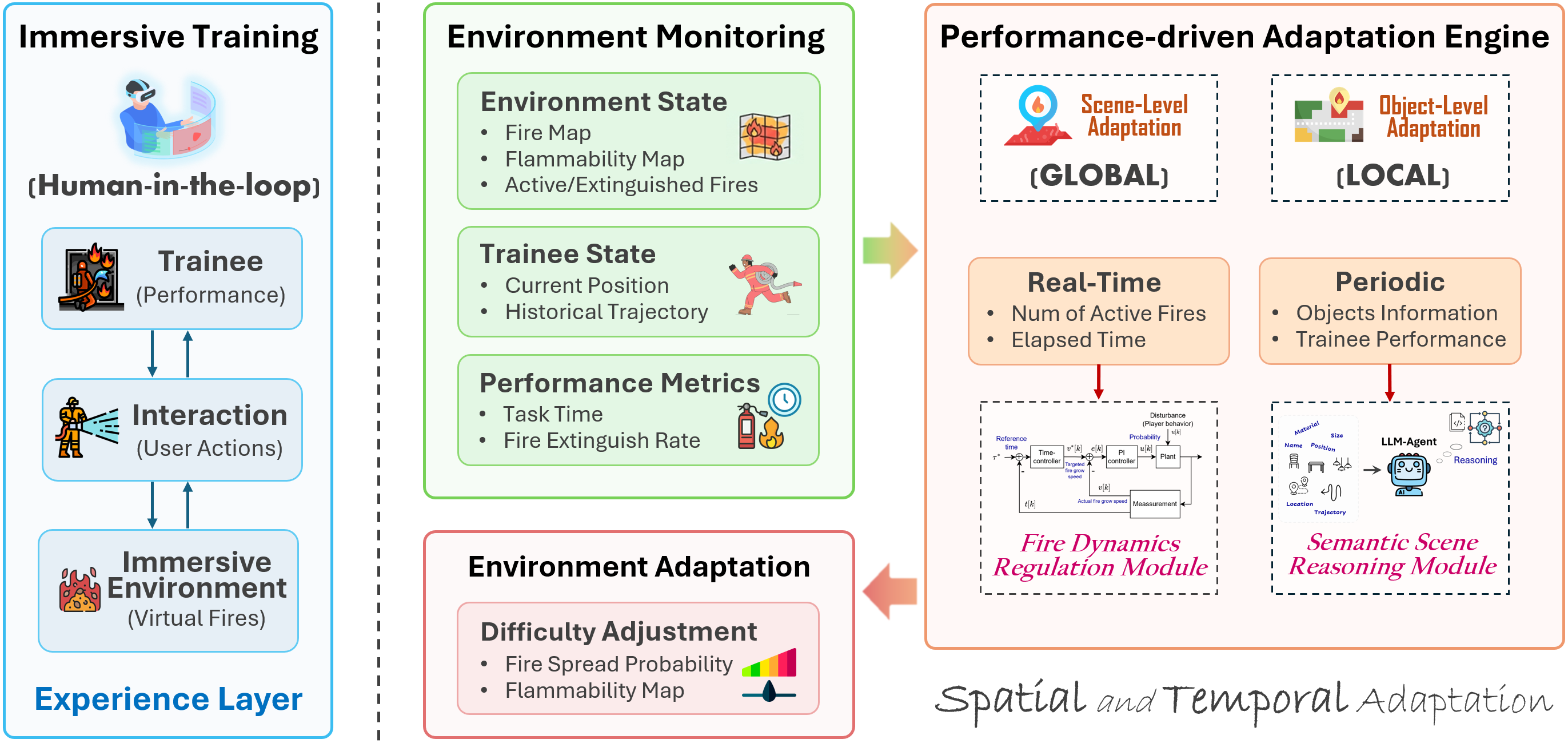}
    \Description{Trainee behavior, environment states, and task performance feed a global fire-spread controller and a local semantic reasoning module, whose updates return to the VR environment.}
    \caption{System overview of ProACT. The system monitors trainee behavior, environment states, and interaction performance during VR firefighter training, and adapts scenario difficulty through a multi-scale adaptive engine. The global module regulates scene-level fire spread probability in real time, while the local module periodically updates object-level flammability based on scene semantics and trainee behavior. These adjustments are fed back to the virtual environment to support continuous personalized training.}
    \label{fig:system-architecture}
    \vspace{-2mm}
\end{figure*}

To address the design challenges identified above, ProACT implements a multi-scale adaptive control architecture that links trainee performance to simulated fire dynamics. As illustrated in Fig.~\ref{fig:system-architecture}, the system continuously monitors three types of signals during the training session: (i) environment states, including the fire map and flammability map; (ii) trainee states, including position and movement trajectory; and (iii) task performance metrics, including elapsed time and fire extinguishing progress. These signals provide a real-time description of how the trainee and fire scenario co-evolve.

Based on these observations, the performance-driven adaptation engine regulates scenario difficulty through two complementary mechanisms: (i) The \textit{Global Fire Dynamics Regulation Module}~(\S~\ref{sec:method:realtime-controller}) continuously adjusts the scene-level fire spread probability according to the gap between the trainee's actual progress and the target completion trajectory, addressing the need for real-time performance-driven regulation (C1); (ii) The \textit{Local Semantic Scene Reasoning Module}~(\S~\ref{sec:method:flammability_map}) periodically updates object-level ignition likelihood by combining scene semantics with recent trainee behavior, supporting fine-grained environment regulation (C2). By coordinating global fire regulation with local semantic updates, ProACT can respond to trainee performance in real time while maintaining stable and plausible fire behavior (C3). The following sections first formalize the environment dynamics underlying ProACT~(\S~\ref{sec:method:environment-dynamics}) and then detail the global and local adaptation modules.

\subsection{Environment Dynamics} \label{sec:method:environment-dynamics}

To support the multi-scale adaptive design introduced above, we represent the VR environment in a form that is both observable and controllable. This representation allows the adaptive engine to track trainee performance and fire evolution in real time while adjusting environment parameters for global fire regulation and local spatial modulation. Specifically, we formulate a lightweight discrete-time abstraction that captures the coupled evolution of fire dynamics, trainee interaction, and adaptive control.

\subsubsection{\textbf{Environment and Trainee Representation}} \label{sec:method:rep}

We model each training session as a discrete-time process indexed by $k = 0,1,2,\dots$, where each step corresponds to a fixed simulation interval $\Delta t$ (e.g., one update cycle of the VR engine). The physical time at step $k$ is given by $t[k] = k\Delta t$, representing the elapsed time since the start of the session.

The 3D scene geometry (e.g., walls, furniture, corridors) is discretized into a bounding grid of size $M_1 \times M_2 \times M_3$, where each cell within this grid represents a spatial region (the unit) susceptible to ignition. This grid representation is chosen because it is simultaneously observable--the adaptive engine can track fire state at each cell --- and controllable, since environment parameters can be adjusted at consistent spatial granularity in real time. At each step $k$, the environment state is characterized by two fields. The \textit{fire map} $s[k] \in \{0,1\}^{M_1 \times M_2 \times M_3}$ encodes which cells are actively burning, with $s_i[k]=1$ indicating that that cell $i$ is actively burning. The \textit{flammability map} $\Phi[k] \in \mathbb{R}_{+}^{M_1 \times M_2 \times M_3}$  encodes the susceptibility to ignition across the scene. For example, wooden surfaces carry higher flammability than concrete. Together, $(s[k], \Phi[k])$ constitute the environment state $x[k]$, which drives fire propagation and serves as the primary input to the adaptive engine.

The trainee interacts with this environment through a behavioral input $w[k]$, which captures how the user physically affects the fire state at step $k$, including spatial position, movement trajectory, and the set of cells recently suppressed. Critically, $w[k]$ is defined at the level of measurable environmental effects rather than inferred user intent. In VR settings, similar motion patterns can arise from different causes, such as deliberate caution, disorientation, or fatigue, making intent inference unreliable. By modeling trainee behavior through observable interaction outcomes, ProACT avoids fragile user modeling while retaining the signals most relevant to fire suppression and hazard evolution.

\subsubsection{\textbf{Stochastic Scenario Evolution}}

In firefighting scenarios, fire spread is uncertain and spatially uneven: flames may propagate to nearby regions depending on local conditions, while trainee suppression actions remove active fires and change which areas remain at risk. We model this interaction between fire growth and trainee intervention as a human-in-the-loop stochastic process, where environment dynamics and user actions jointly determine how the scenario evolves over time.

To make this process tractable for real-time adaptation, we describe the VR environment using a controlled stochastic state representation. The environment state at step $k$ is given by $x[k] = (s[k], \Phi[k])$, where the fire map $s[k]$ represents the spatial configuration of active flames and the flammability field $\Phi[k]$ encodes the ignition potential across the scene. The system evolves under two distinct driving factors: a global control signal $u[k] \in [0,1]$ generated by the adaptive engine, and an exogenous interaction signal $w[k]$ induced by the trainee's real-time trajectory and suppression actions. Accordingly, the next state is governed by the transition law
\begin{equation} \label{eqn:state_transition}
    x[k+1] \sim P\big(\,\cdot \mid x[k], u[k], w[k]\big),
\end{equation}
where $P$ is the stochastic kernel over the environment state space. At the beginning of each session, multiple ignition seeds are randomly initialized across combustible regions of the scene.

We constrain fire propagation to be spatially contiguous. Specifically, an inactive cell can ignite at the next step only if it lies within the 26-neighborhood of a currently burning cell, where the 26-neighborhood refers to the surrounding cells in a 3D “$3 \times 3 \times 3$” grid excluding the center cell. The control signal $u[k]$ controls the base likelihood of propagation: a higher value increases scene-level fire spread, while a lower value slows it down. The flammability field $\Phi[k]$ further modulates ignition locally, introducing spatial heterogeneity based on object- and region-level properties. In parallel, the behavioral input $w[k]$ captures user-induced suppression by removing burning cells from the fire map, thereby reducing subsequent propagation in affected regions.


\begin{figure*}[ht]
    \centering
    \includegraphics[width=.8\linewidth]{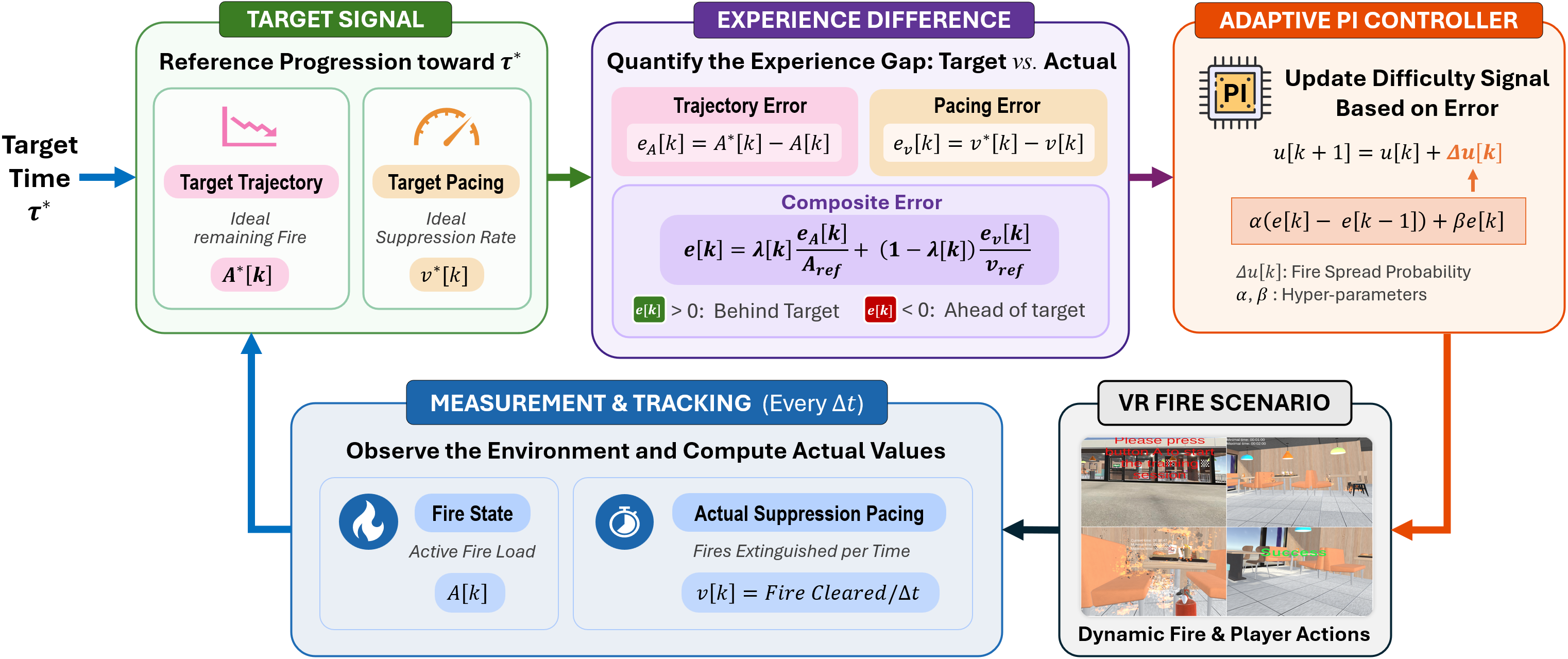}
    \Description{A feedback loop compares observed fire suppression with target progression and adjusts the global fire-spread probability.}
    \caption{System flow of the Global Fire Dynamics Regulation Module. The controller continuously measures the experience gap between the trainee's actual progress and the target progression, and dynamically tunes the scene-level fire spread probability to avoid both overload and under-challenge.}
    \label{fig:RealTimeAdaptation}
\end{figure*}

\subsubsection{\textbf{Regulation Goal}}

The objective of ProACT is not to minimize completion time, but to keep the training scenario within a target challenge range, where the task is neither completed too quickly nor becomes too difficult to finish.

To capture this goal, we define a physical target completion time $\tau^*$ and an acceptable completion-time window $[\underline{\tau}, \overline{\tau}]$, where $\underline{\tau} < \tau^* < \overline{\tau}$. 

Let $A[k] = \sum_{i}s_i[k]$ denote the active fire load at step $k$. We define the completion step as the first update step at which no active fire remains:
\[
    k_c := \inf\{k \ge 0 : A[k] = 0\}.
\]
Then, the corresponding completion time is $\tau_c := t[k_c] = k_c\Delta t$.  

A training session is considered \textit{successful} (on-target) if its completion time falls within the target window, i.e., $\tau_c \in [\underline{\tau}, \overline{\tau}]$. Conversely, 
the session is a \emph{failure} if $\tau_c > \overline{\tau}$, indicating the trainee is over-challenging, or $\tau_c < \underline{\tau}$, the scenario is considered under-challenging. 

We can therefore formalize this high-level regulation objective as a human-in-the-loop stochastic control problem. Given the evolving environment state~$\{x[k]\}_{k \geq 1}$ and exogenous trainee interaction~$\{w[k]\}_{k \geq 1}$, the goal is to select the global spread probability and local flammability field over time to maximize the likelihood of on-target completion:
\[
        \max_{\{u[k], \Phi[k]\}_{k \geq 1}} \; \mathbb{P}\left(\tau_c \in [\underline{\tau}, \overline{\tau}]\right), \quad \textrm{subject to stochastic dynamics} \; \eqref{eqn:state_transition}.
\]

Because the trainee's actions $w[k]$ are highly variable and non-deterministic, solving this problem exactly is intractable with high computation. Instead, ProACT adopts a tractable multi-scale control strategy. We decouple the adaptation into two complementary mechanisms: a fast global control loop that continuously adjusts the scene-level spread probability $u[k]$ based on real-time trajectory tracking (detailed in~\S~\ref{sec:method:realtime-controller}), and a slower local reasoning loop that periodically updates the spatial flammability field $\Phi[k]$ based on semantic and behavioral context (detailed in~\S~\ref{sec:method:flammability_map}). Together, these loops coordinate real-time difficulty regulation with semantically grounded local adaptation, steering the scenario toward the target challenge window.


\subsection{Global Fire Dynamics Regulation Module}
\label{sec:method:realtime-controller}

The \textit{Global Fire Dynamics Regulation Module} instantiates ProACT's real-time feedback controller for regulating scene-level fire progression based on trainee performance. Rather than relying on predefined difficulty levels, the controller continuously adjusts the global fire spread probability so that the evolving environment remains aligned with the target challenge range. As shown in Fig.~\ref{fig:RealTimeAdaptation}, this design enables real-time adaptation while preserving the realism and continuity of the training experience.

\subsubsection{\textbf{Defining the Target Pacing}} \label{sec:method:target_pace}

To regulate scenario progression, we translate the target completion time $\tau^*$ into two reference signals: a target pacing signal and a target trajectory for the remaining fire load. The pacing signal captures the suppression rate required at the current step, while the target trajectory provides a reference for the overall evolution of the session.

Recall that $A[k] = \sum_i s_i[k]$ represents the number of grid cells currently burning. Given the elapsed time $t[k]$, the remaining time budget is $\tau^* - t[k]$. We define the target pacing signal as
\begin{equation}
    v^*[k] := \frac{A[k]}{\tau^* - t[k]},
    \qquad t[k]<\tau^*,
\end{equation}
which represents the average suppression rate required from the current step onward to extinguish all remaining fire by $\tau^*$. This pacing signal is recomputed from the current fire load and remaining time, providing an instantaneous reference for the controller.

To complement it, we introduce a target trajectory for the remaining fire load, denoted by $A^*[k]$, which serves as a smoothed reference for how the fire load should decrease under ideal progression toward completion. This trajectory is initialized as $A^*[0] = A[0]$ and evolves recursively as
\begin{equation}
    A^*[k+1] = \max\left\{0, \ A^*[k] - \Delta t\cdot \frac{A^*[k]}{\tau^*-t[k]} \right\}.
\end{equation}

This construction ensures that $A^*[k]$ decreases smoothly and monotonically to zero at the target time, providing a stable reference that is independent of the stochastic fire dynamics. Importantly, $A^*[k]$ is not intended to model the true evolution of the fire, but rather to serve as a control-oriented baseline for assessing progression. Together, $v^*[k]$ and $A^*[k]$ provide complementary references: the former captures the required suppression pace from the current state, while the latter encodes the desired overall trajectory of the task.

\subsubsection{\textbf{Measuring the Experience Gap}}

To evaluate whether the session is progressing as intended, the system constructs an observed pacing signal from real-time fire-state measurements. We first define the actual suppression pacing as the change in active fire load over one update interval
\[
    v[k] := \frac{A[k] - A[k+1]}{\Delta t}.
\]

This quantity captures the net reduction in active fire load induced by both trainee suppression and stochastic fire evolution.

Based on the reference signals $v^*[k]$ and $A^*[k]$, we define two error components. The trajectory error $e_A[k] := A[k] - A^*[k]$ captures whether the remaining fire load is above or below the desired trajectory. The pacing error $e_v[k] := v^*[k] - v[k]$ captures whether the observed suppression rate is slower or faster than the rate required to reach the target time. To integrate these two aspects into a unified feedback signal, ProACT normalizes the trajectory and pacing errors by reference scales $A_{\textrm{ref}}$ and $v_{\textrm{ref}}$, respectively, and computes a weighted experience gap: 
\begin{equation}
    e[k] = \lambda[k]\frac{e_A[k]}{A_{\textrm{ref}}} + (1-\lambda[k]) \frac{e_v[k]}{v_{\textrm{ref}}},
\end{equation}
where $0 \leq \lambda[k] \leq 1, \forall k=1,2,3,\ldots$ controls the relative weighting between the two normalized error terms. The two errors capture complementary aspects of the training experience. The pacing error $e_v[k]$ provides a short-term responsiveness signal: it reflects whether the trainee is currently suppressing fire faster or slower than the rate required to reach the target window. This is especially useful early in the session, when the trainee may begin to gain confidence, learn the spatial layout, or improve their handling of the extinguisher. If the observed suppression pace becomes faster than expected, this term allows the controller to respond by increasing the fire-spread probability, keeping the task appropriately challenging rather than allowing it to become too easy.

In contrast, the trajectory error $e_A[k]$ reflects the task-level objective: it measures whether the remaining active fire load is above or below the desired progression. As the session approaches the target completion window, this term becomes increasingly important because the controller should focus less on moment-to-moment behavioral fluctuations and more on whether the trainee is on track to complete the task at the intended time. Therefore, in our implementation, the weighting is time-adaptive: early in the session, the controller gives more weight to $e_v[k]$ to rapidly adapt to the trainee's observed behavior; later in the session, the weight gradually shifts toward $e_A[k]$, prioritizing convergence to the desired fire-load trajectory and supporting successful task completion within the target window.

A positive value of $e[k]$ indicates that the trainee is falling behind the required progression, while a negative value indicates that the trainee is progressing faster than required. This formulation enables the controller to account for both short-term responsiveness and long-term task progression. Importantly, it avoids explicitly estimating user skill or intent. Instead, it adapts directly from task-level measurements, making the system robust to variability in trainee behavior and stochastic fire dynamics.

\subsubsection{\textbf{Real-Time Regulation}}

To regulate environment evolution, ProACT updates the global spread probability $u[k]$ based on the experience gap $e[k]$. The objective of this update is to continuously steer the system toward the experience-balanced state defined by $e[k] = 0$, where the observed progression aligns with the target progression.

ProACT adopts a feedback structure inspired by incremental proportional–integral (PI) control, adapted to operate on task-level progression signals. Specifically, the update rule is given by
\begin{align}
    \Delta u[k] &= \alpha\big(e[k] - e[k-1]\big) + \beta e[k],   \nonumber \\
    u[k+1] &= \textrm{clip}(u[k] - \Delta u[k], \, 0, \, 1),
\end{align}
where $\alpha$ and $\beta$ are tunable control parameters. The clip operation ensures that $u[k+1]$ is a valid probability. With this sign convention, positive gaps reduce the spread probability to alleviate overload, while negative gaps increase it to preserve challenge.

This update is designed to satisfy two complementary requirements. First, the controller should react quickly when the trainee's pacing changes abruptly. The term $\alpha(e[k] - e[k-1])$ captures the step-to-step change in the experience gap and enables rapid adjustment to sudden performance shifts. For example, if the trainee quickly clears a large region of fire, the gap becomes more negative, causing the controller to increase the spread probability and preserve challenge. Second, the controller should correct sustained mismatch without causing abrupt or unrealistic changes in difficulty. Because $\beta e[k]$ is applied incrementally to $u[k]$, persistent deviations accumulate over time and drive sustained correction between the target and observed progression.

\begin{figure*}[t]
    \centering
    \includegraphics[width=.8\linewidth]{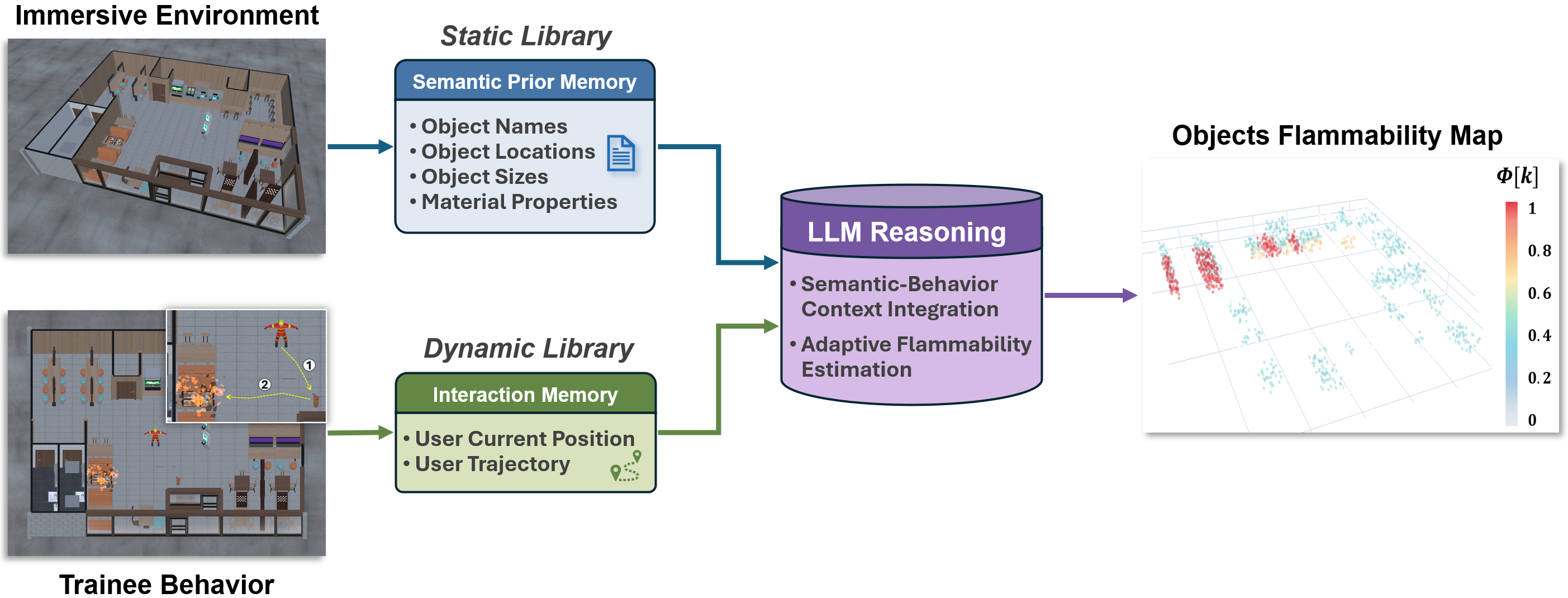}
    \Description{Static object metadata and recent trainee context feed an LLM reasoning agent that selects regions for bounded flammability updates.}
    \caption{Overview of the Local Semantic Scene Reasoning Module. The module combines static scene semantics, including object name, location, size, and material category, with dynamic trainee context, including current position and recent movement trajectory. An LLM-based reasoning agent uses these inputs to guide localized updates to the object-level flammability map for local environment regulation.}
    \label{fig:llm-env}
    \vspace{-2mm}
\end{figure*}

This feedback mechanism provides continuous and smooth adjustment of fire dynamics without abrupt transitions between predefined difficulty levels. In contrast to static curricula or discrete difficulty tiers, the controller operates directly on real-time progression signals, enabling individualized adaptation without explicitly modeling user skill.


\subsection{Local Semantic Scene Reasoning Module}
\label{sec:method:flammability_map}

While the global controller regulates the overall pace of fire progression by adjusting the scene-level spread probability~$u[k]$, effective training also requires local adaptation that accounts for spatial differences in object semantics and trainee interaction patterns. To address this need, ProACT introduces the \textit{Local Semantic Scene Reasoning Module} that periodically updates the flammability field based on scene semantics and recent trainee behavior. As shown in Fig.~\ref{fig:llm-env}, the module consists of a Static Library that encodes persistent object-level properties and a Dynamic Library that summarizes recent spatial behavioral patterns from the trainee's trajectory. An LLM-based reasoning agent fuses these two information sources to guide localized, context-aware updates to the flammability field.

\subsubsection{\textbf{Static Library: Semantic Prior Memory}}

The Static Library establishes a persistent semantic prior over the virtual scene and initializes a spatially resolved baseline flammability field. For each object in the environment, we extract low-level metadata $m$ from the Unity scene at initialization, including object location, size, scale, and material descriptors. Because asset metadata can be heterogeneous or overly specific, ProACT uses an LLM-assisted inference step to map each object's metadata into a standardized material category $c$ (e.g., wood, textile, plastic, metal), denoted as $c = f_{\textrm{{\tiny{LLM}}}}(m)$. Each category is associated with a baseline flammability coefficient $\varphi(c) \in [0,1]$, which captures the relative ignition tendency of that material class. The classified scene is then discretized onto the spatial grid introduced in Section~\ref{sec:method:rep}. Object volumes are projected onto this grid, and each occupied cell $j$ is assigned an initial flammability value according to its semantic class, i.e., $\Phi_j[0] = \varphi(c)$. Cells not occupied by combustible objects are assigned a near-zero baseline value.

If multiple objects overlap within a single cell, we assign the highest flammability coefficient among the occupants. The resulting field $\Phi[0]$ serves as the persistent prior over the environment and constitutes the initial condition for the adaptive flammability field. Besides, the Static Library remains fixed throughout the training session. All subsequent adaptations are therefore applied as deviations from this semantically grounded baseline, ensuring that environment modifications remain physically interpretable.

\subsubsection{\textbf{Dynamic Library: Interaction Memory}} 

The Dynamic Library captures short-term trainee-environment interaction during the session. Unlike the Static Library, which encodes persistent scene properties, the Dynamic Library is transient and updated periodically at intervals of $\Delta t_k$ (e.g., every 10 seconds). At each update, it summarizes the trainee's current position and recent movement trajectory. These signals provide a compact description of where the trainee has recently operated, where fire pressure remains high, and which nearby regions are available for controlled local adaptation. The Dynamic Library does not directly determine the flammability update. Instead, it serves as a behavioral context for the reasoning module. This separation allows ProACT to use measurable interaction signals without making brittle assumptions about user intent.

\begin{figure*}[t]
\centering
\includegraphics[width=.7\linewidth]{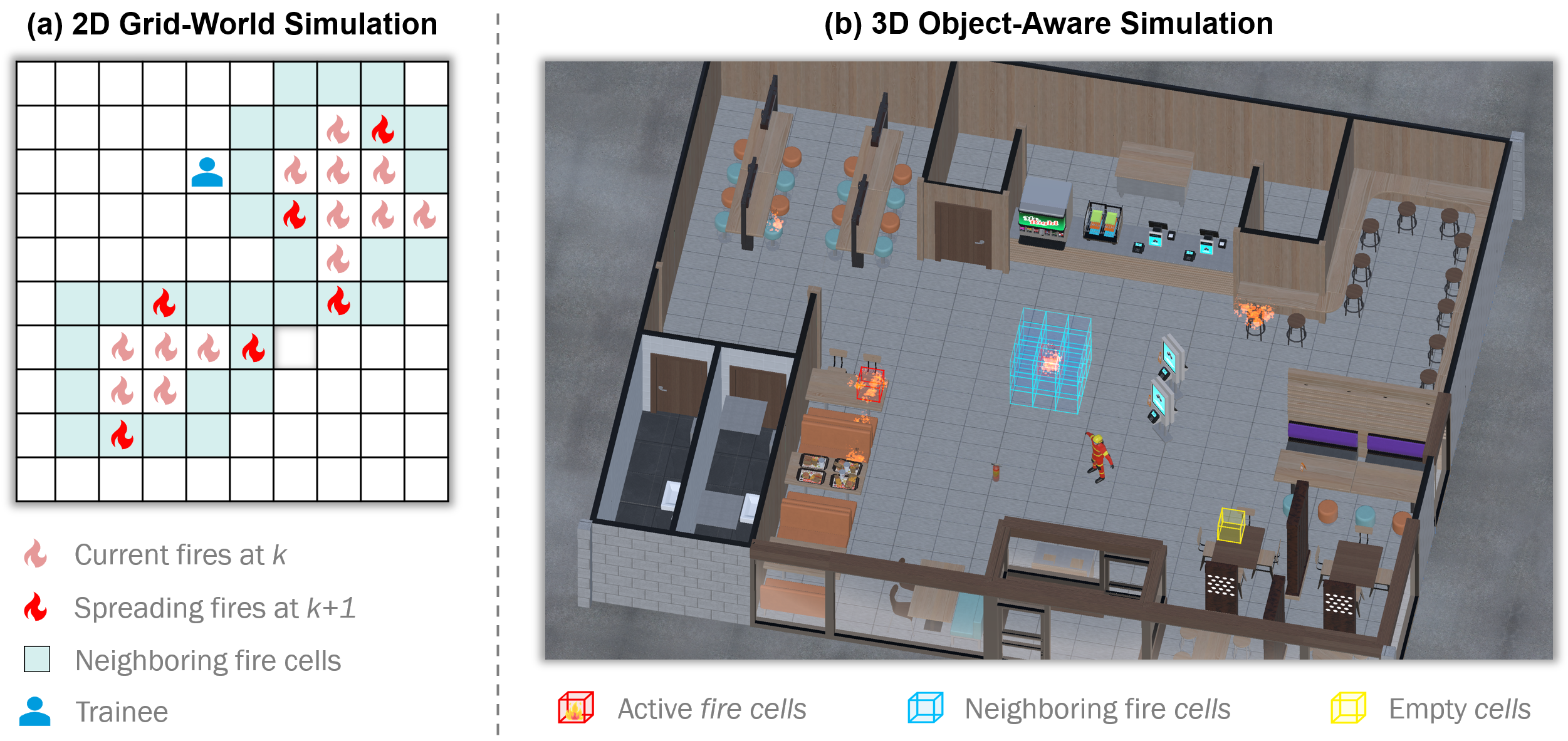}
    \Description{Two simulation settings: a homogeneous 2D fire grid with a simulated trainee and a 3D scene containing objects with heterogeneous flammability.}
\vspace{-3mm}
\caption{Simulation environments used for the staged evaluation. (a)~The 2D grid-world abstracts fire propagation and trainee suppression on a homogeneous
grid, isolating the effect of global challenge regulation; current fire cells at step $k$, candidate spreading cells at $k+1$, neighboring fire cells, and the simulated trainee are visualized. (b)~The 3D object-aware simulation introduces scene geometry, object-level structure, and heterogeneous flammability,
providing a more realistic setting for evaluating the coordination of global fire regulation and local semantic adaptation.}
\label{fig:simsettings}
\vspace{-2mm}
\end{figure*}

\subsubsection{\textbf{Context-Aware Fusion via LLM}}

At each local update interval, ProACT constructs a compact description of the current local scene using the Static Library and Dynamic Library. The Static Library provides persistent scene semantics, including object categories, material priors, spatial locations, and the baseline flammability map $\Phi[0]$. The Dynamic Library complements this scene-level prior with recent trainee behavior, nearby fire pressure, trajectory information, and task-level progress signals such as the current experience gap and estimated completion status.

Given this summary, the LLM identifies semantically appropriate regions where local flammability adjustments can be applied. The prompt includes a small set of instruction examples that describe desirable local adaptations under different progress conditions, such as increasing flammability when the session is under-challenging or reducing flammability when the trainee appears overloaded. At runtime, the LLM uses these instructions and the current scene summary to select semantically appropriate regions for local adjustment. In this way, the LLM incorporates scene semantics and behavioral context into local adaptation without directly controlling global fire dynamics.

To maintain stability, the LLM-guided output is applied only as a bounded local modification to the flammability map. The local flammability field is updated periodically every $n$ steps and the update direction and magnitude are determined by target-window progress signals, while the LLM-selected region determines where the adjustment is applied. The resulting update is applied as
\begin{equation}
    \quad \Phi[n(k+1)] =
    \mathrm{clip}\left(
    \Phi[nk] + \Delta \Phi_{\mathrm{loc}}[nk],
    \Phi_{\min}, \Phi_{\max}
    \right),
\end{equation}
where $\Delta \Phi_{\mathrm{loc}}[nk]$ denotes the bounded local flammability update guided by the semantic reasoning module. The clipping operation ensures that all flammability values remain within the valid range $[\Phi_{\min}, \Phi_{\max}]$. Before applying the update, ProACT performs a lightweight consistency check that enforces valid target regions, bounded update magnitude, and consistency with the current progress state. These constraints keep local semantic adaptation aligned with the global challenge-regulation objective while preventing abrupt changes to the simulated fire behavior.


\section{Experimental Setup}

\subsection{Implementation Details}
ProACT was implemented in Unity 2022.3.43f1 and deployed on a Meta Quest~3 head-mounted display for the VR firefighter training system used in the user
study. For the simulation-based evaluation, we developed a Python simulator based on the same fire-dynamics abstraction and adaptation logic described in
Section~\ref{sec:method:environment-dynamics}. Across both implementations we use the discrete update index $k$ as the common time base, with elapsed time $t[k]=k\Delta t$. In the Unity
implementation, the global controller runs every $0.5$\,s ($\Delta t=0.5$\,s), so completion time is reported in seconds; in simulation we use a normalized time scale with $\Delta t=1$, so completion time is reported in steps. The same control logic is used in both settings; one step simply represents a different physical duration. The Local Semantic Scene Reasoning Module runs at a slower time scale and updates the object-level flammability map every $20$\,s (steps), using GPT-5-mini~\cite{singh2025openai} to reason over scene semantics and trainee behavior and to guide bounded flammability adjustments in selected local regions. All simulations were run on a Windows desktop with an Intel Core i9-14900KF CPU, 64\,GB RAM, and an NVIDIA GeForce RTX 4070 GPU.

\subsection{Baselines}
Across the study, we compare ProACT with three baseline policies. \emph{StaticEasy} and \emph{StaticHard} use fixed spread probabilities throughout
the session: StaticEasy uses a lower controller probability for fire spreading and StaticHard a higher one, representing under-challenging and over-challenging static conditions, respectively. Neither adapts to the trainee's behavior or progress. \emph{Reactive} is a lightweight feedback baseline inspired by dynamic difficulty adjustment in games~\cite{sepulveda2019exploring}. It adjusts the fire spread probability online using coarse task-progress signals: at each step it estimates the trainee's recent extinguishing rate with an exponential moving average, predicts the expected completion time from the current active fire count, and compares it with the target window. If completion is predicted to be too early, it increases the spread probability to preserve challenge; if too late, it decreases the spread probability to reduce overload. Unlike ProACT, Reactive uses neither target-trajectory tracking, semantic scene understanding, nor LLM-based flammability adaptation.

\begin{table*}[t]
\centering
\caption{On-target completion rate and completion-time statistics in the \textbf{2D grid-world} simulation. OT denotes the on-target completion rate (\% of
episodes completed within the target window); Mean and Std denote the mean and standard deviation of completion time across 100 independent episodes.
\textbf{Bold}: highest OT for each capability level; \underline{underline}: second-highest OT.}
\label{tab:2d_summary}
\renewcommand{\arraystretch}{1.1}
\setlength{\tabcolsep}{5pt}

\resizebox{.7\linewidth}{!}{%
\begin{tabular}{l|ccc|ccc|ccc|ccc}
\toprule
\multirow{2}{*}{Method}
& \multicolumn{3}{c|}{$\epsilon=0.05$}
& \multicolumn{3}{c|}{$\epsilon=0.20$}
& \multicolumn{3}{c|}{$\epsilon=0.40$}
& \multicolumn{3}{c}{$\epsilon=0.60$} \\
\cmidrule(lr){2-4}\cmidrule(lr){5-7}\cmidrule(lr){8-10}\cmidrule(lr){11-13}
& OT & Mean & Std & OT & Mean & Std & OT & Mean & Std & OT & Mean & Std \\
\midrule
StaticEasy & 2.0 & 34.83 & 43.50 & 2.0 & 73.91 & 62.78 & 1.0 & 110.91 & 58.03 & 0.0 & 143.36 & 30.47 \\
StaticHard & 0.0 & 135.90 & 42.97 & 0.0 & 140.29 & 36.45 & 0.0 & 149.63 & 13.63 & 0.0 & 151.00 & 0.00 \\
Reactive   & \underline{54.0} & 63.22 & 16.87 & \underline{74.0} & 72.67 & 16.43 & \underline{75.0} & 95.63 & 21.81 & \underline{31.0} & 124.75 & 28.62 \\ \midrule
ProACT     & \textbf{77.0} & 73.17 & 15.07 & \textbf{90.0} & 83.29 & 19.47 & \textbf{78.0} & 95.40 & 25.18 & \textbf{60.0} & 108.46 & 28.86 \\
\bottomrule
\end{tabular}%
}
\vspace{-3mm}
\end{table*}

\begin{figure*}[t]
    \centering
    \includegraphics[width=\linewidth]{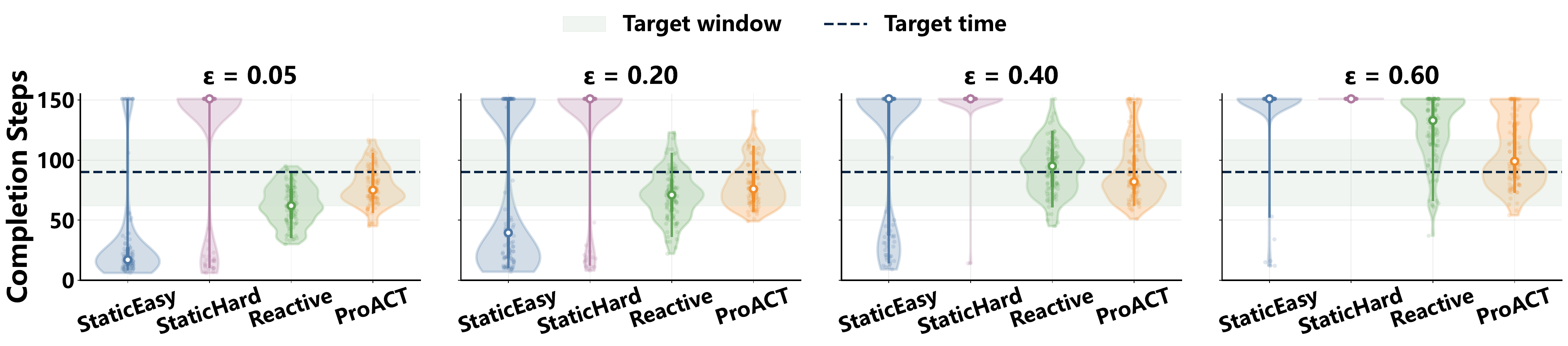}
    \Description{Completion-time distributions for four methods at four simulated capability levels in the 2D grid-world, with the target completion window shaded.}
    \caption{Completion-time distributions in the 2D grid-world simulation across the four capability levels ($\epsilon=0.05,0.20,0.40,0.60$; larger $\epsilon$ = less consistent behavior). Each panel aggregates 100 episodes; the shaded band marks the target window $[\underline{\tau},\overline{\tau}]$ and the dashed line the target time $\tau^*$. ProACT achieves the highest on-target completion rate across all evaluated capability levels and concentrates more completions within the target window than the baselines.}
    \label{fig:dist2d}
\end{figure*}

\section{Simulated Environment Evaluation}
Before the VR user study, we evaluate ProACT through controlled simulations to examine its adaptive behavior under repeatable conditions. The simulation study answers two questions: whether the Global Fire Dynamics Regulation Module can steer completion time toward the target challenge window, and whether ProACT can coordinate global fire regulation with local semantic adaptation when fire dynamics are shaped by spatial layout, object structure, and heterogeneous flammability. As shown in Figure~\ref{fig:simsettings}, we use a staged evaluation with two environments. The 2D grid-world removes object semantics and material differences, enabling a focused analysis of global challenge regulation. The 3D object-aware simulation introduces scene geometry and semantic flammability, allowing us to assess the combined behavior of the global and local modules in a setting closer to the immersive training environment.

\begin{figure*}[t]
    \centering
    \includegraphics[width=1\linewidth]{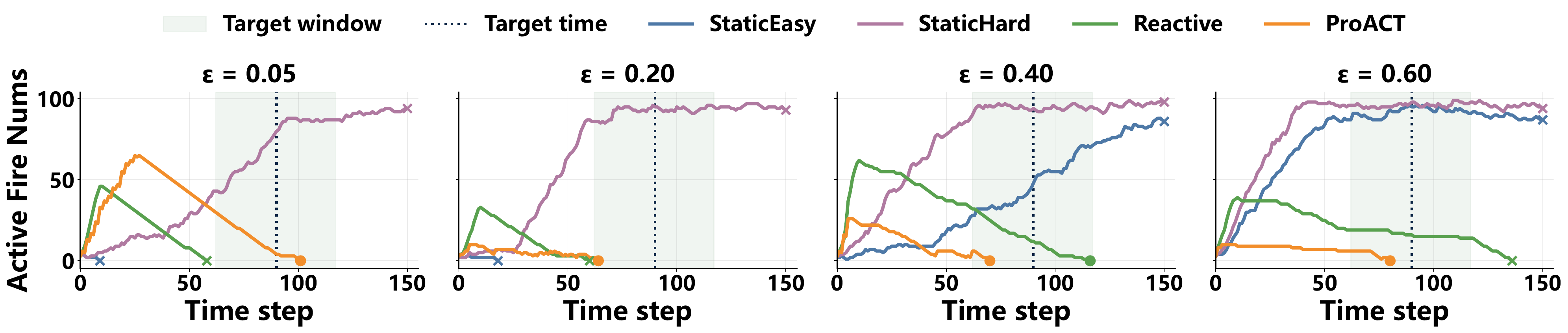}
    \Description{Matched 2D fire-load trajectories for four methods and four capability levels, with markers indicating completion outcomes relative to the target window.}
    \caption{Representative fire-load trajectories in the 2D grid-world simulation. Each panel shows a matched episode across StaticEasy, StaticHard, Reactive, and ProACT under one simulated trainee capability level, with active fire load $A[k]$ plotted over time using a linear y-axis. The shaded region denotes the target challenge window $[\underline{\tau}, \overline{\tau}]$, and the vertical dashed line marks the target time $\tau^*$. Markers denote episode outcomes: circles indicate on-target completion, while crosses indicate off-target completion or termination at the 150-step horizon.}
    \label{fig:2d_completion_nums}
\end{figure*}

\subsection{Evaluation Setup}
Across both simulations, we evaluate the same four methods: StaticEasy, StaticHard, Reactive, and ProACT. To emulate simulated trainees of different capabilities, we use an $\epsilon$-greedy behavior model: at each step the trainee moves toward the nearest active fire with probability $1-\epsilon$ and takes a random exploratory action with probability $\epsilon$. Here $\epsilon$ is a proxy for capability---lower $\epsilon$ is more consistent, effective suppression behavior, higher $\epsilon$ is less consistent---and we evaluate $\epsilon\in\{0.05,0.20,0.40,0.60\}$. Following the regulation goal in Section~\ref{sec:method:environment-dynamics}, the objective is not to minimize completion time but to complete the task within the target window. With $\Delta t=1$, we set the target completion time to $\tau^*=90$ and define $[\underline{\tau},\overline{\tau}]=[(1-\rho)\tau^*,(1+\rho)\tau^*]$ with $\rho=0.3$, yielding a target window of $[63,117]$ steps. For each method and $\epsilon$ we run 100 independent episodes with matched random seeds across methods, and report the on-target completion rate (OT), mean and standard deviation of completion time.

\subsection{2D Grid-World Simulation}
We first evaluate the Global Fire Dynamics Regulation Module in the 2D grid-world (Figure~\ref{fig:simsettings}a). This setting removes 3D geometry, object semantics, and material differences, so the flammability field is fixed at $\Phi[k]=1$ for all cells, isolating global challenge regulation. The environment is a $10\times10$ grid with binary fire states. Each episode starts with three randomly initialized seed fires and evolves with $\Delta t=1$ until all fires are extinguished or a 150-step horizon is reached. At each step, fire may spread to the 8-neighborhood while the simulated trainee suppresses one local region.

\begin{table*}[t]
\centering
\caption{On-target completion rate and completion-time statistics in the \textbf{3D object-aware} simulation. OT denotes the on-target completion rate (\% of episodes completed within the target window); Mean and Std denote the mean and standard deviation of completion time across 100 independent episodes.
\textbf{Bold}: highest OT for each capability level; \underline{underline}: second-highest OT.}
\label{tab:3d_summary}
\renewcommand{\arraystretch}{1.1}
\setlength{\tabcolsep}{5pt}

\resizebox{.7\linewidth}{!}{%
\begin{tabular}{l|ccc|ccc|ccc|ccc}
\toprule
\multirow{2}{*}{Method}
& \multicolumn{3}{c|}{$\epsilon=0.05$}
& \multicolumn{3}{c|}{$\epsilon=0.20$}
& \multicolumn{3}{c|}{$\epsilon=0.40$}
& \multicolumn{3}{c}{$\epsilon=0.60$} \\
\cmidrule(lr){2-4}\cmidrule(lr){5-7}\cmidrule(lr){8-10}\cmidrule(lr){11-13}
& OT & Mean & Std & OT & Mean & Std & OT & Mean & Std & OT & Mean & Std \\
\midrule
StaticEasy & 16.0 & 71.77 & 42.04 & \underline{40.0} & 80.14 & 40.43 & \underline{65.0} & 98.52 & 34.09 & \underline{23.0} & 134.53 & 23.72 \\
StaticHard & \underline{41.0} & 103.95 & 42.26 & 35.0 & 119.38 & 37.30 & 12.0 & 141.13 & 22.39 & 1.0 & 150.31 & 6.87 \\
Reactive   & 22.0 & 76.08 & 43.23 & \underline{40.0} & 84.14 & 41.12 & \textbf{66.0} & 98.92 & 33.67 & \underline{23.0} & 133.97 & 24.47 \\ \midrule
ProACT     & \textbf{49.0} & 108.34 & 37.41 & \textbf{48.0} & 107.58 & 37.78 & 61.0 & 93.66 & 35.66 & \textbf{70.0} & 98.10 & 30.88 \\
\bottomrule
\end{tabular}
}
\end{table*}

Table~\ref{tab:2d_summary} and Figure~\ref{fig:dist2d} summarize the completion-time distributions across capability levels. The fixed baselines illustrate the
difficulty of selecting a single static spread setting. \emph{StaticEasy} often produces under-challenging episodes for consistent trainees: at $\epsilon=0.05$ its mean completion time is only $34.83$ steps---far earlier than $\tau^*=90$---%
with an OT of $2.0\%$. As capability decreases, its completion times shift toward the horizon and become highly variable, rising from $73.91$ steps at
$\epsilon=0.20$ to $143.36$ at $\epsilon=0.60$. \emph{StaticHard} exhibits the opposite failure mode, with mean completion times close to the horizon under
lower capability and near-zero OT across all $\epsilon$. \emph{Reactive} improves over the fixed baselines by adjusting difficulty online, reaching OT of
$54.0/74.0/75.0\%$ for $\epsilon=0.05/0.20/0.40$, but degrades to $31.0\%$ at $\epsilon=0.60$. In contrast, \emph{ProACT} achieves the highest OT at every
capability level ($77.0/90.0/78.0/60.0\%$) and keeps the mean completion time closer to the window ($73.2/83.3/95.4/108.5$ steps), with completions more
concentrated around the shaded target window than the baselines. These results indicate that the global feedback controller adapts the fire-spread probability across capability levels and guides completion times toward the target window.

Fig.~\ref{fig:2d_completion_nums} further explains these distribution-level results through representative matched fire-load trajectories. The fixed baselines show two opposite failure modes: \textit{StaticEasy} often drives the active fire load $A[k]$ to zero before the target window, making the scenario under-challenging for more consistent trainees, whereas \textit{StaticHard} keeps fire load high and leads to overload. \textit{Reactive} partially reduces these extremes, but its threshold-based updates can lag behind the evolving fire state, causing trajectories to finish too early or drift outside the target window. In contrast, \textit{ProACT} better aligns the reduction of $A[k]$ with the target challenge window. Rather than minimizing fire load as quickly as possible, it allows fire dynamics to evolve and then gradually guides completion toward the target time. This pattern is consistent across capability levels: ProACT avoids premature completion at low $\epsilon$ and maintains near-target completion at high $\epsilon$, showing more stable training-pace regulation.

\begin{figure*}[t]
    \centering
    \includegraphics[width=\linewidth]{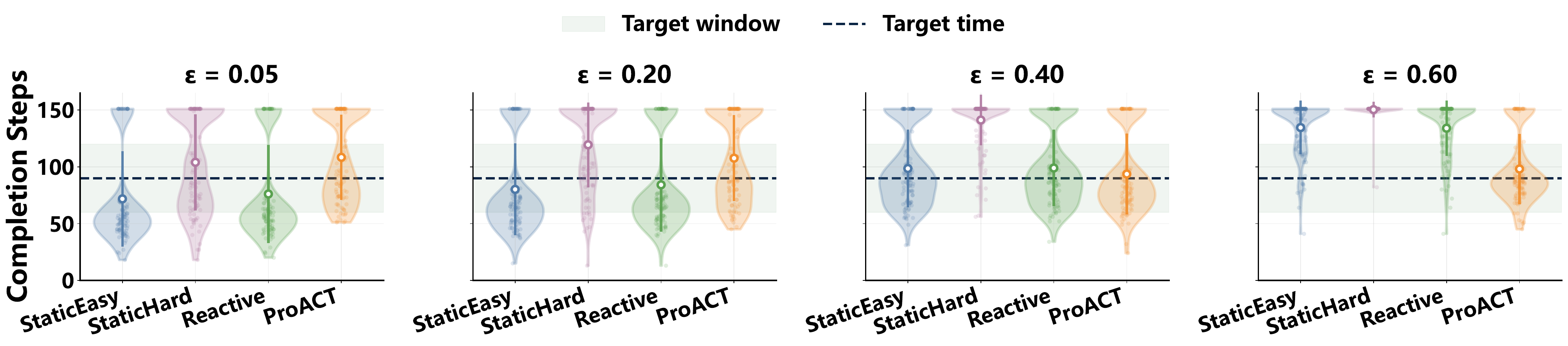}
    \Description{Completion-time distributions for four methods at four simulated capability levels in the 3D object-aware environment, with the target window shaded.}
    \caption{Completion-time distributions in the 3D object-aware simulation across the four capability levels ($\epsilon=0.05,0.20,0.40,0.60$; larger
    $\epsilon$ = less consistent behavior). Each panel aggregates 100 episodes; the shaded band marks the target window and the dashed line the target time.
    ProACT keeps completions inside the window more consistently than StaticEasy, StaticHard, and Reactive.}
    \label{fig:dist3d}
\end{figure*}

\begin{figure*}[t]
    \centering
    \includegraphics[width=1\linewidth]{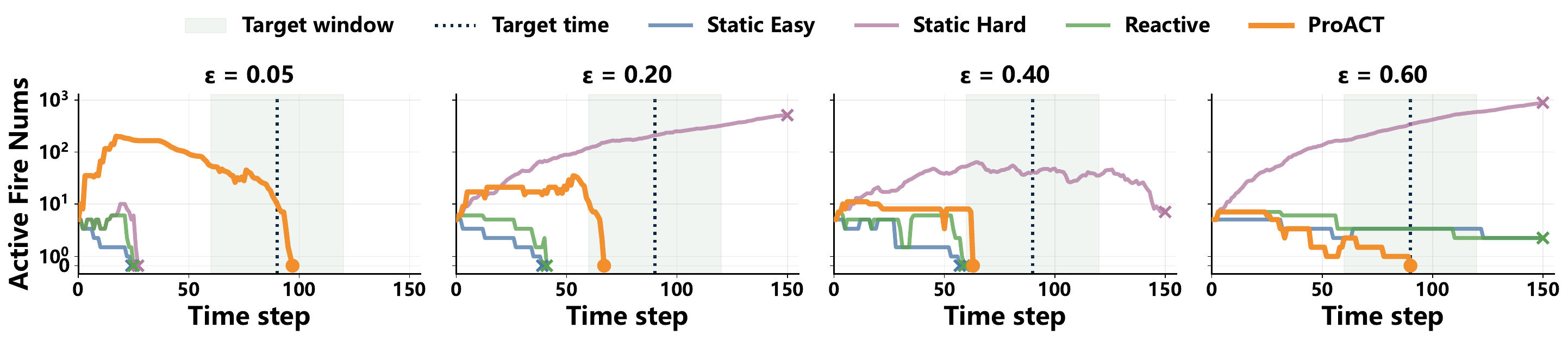}
    \Description{Matched 3D fire-load trajectories for four methods and four capability levels on a symmetric logarithmic scale, with completion outcomes marked.}
    \caption{Representative fire-load trajectories in the 3D object-aware simulation. Each panel shows a matched episode across StaticEasy, StaticHard, Reactive, and ProACT under one simulated trainee capability level, with active fire load $A[k]$ plotted over time using a symmetric logarithmic y-axis. The shaded region denotes the target challenge window $[\underline{\tau}, \overline{\tau}]$, and the vertical dashed line marks the target time $\tau^*$. Markers denote episode outcomes: circles indicate on-target completion, while crosses indicate off-target completion or termination at the 150-step horizon.}
    \label{fig:3d_completion_nums}
\end{figure*}

\subsection{3D Object-Aware Simulation}
We next evaluate ProACT in the 3D object-aware simulation (Figure~\ref{fig:simsettings}b), which introduces scene geometry, object-level structure, and heterogeneous flammability. Unlike the 2D grid-world, where all cells share the same flammability, propagation here is affected by both the global spread probability and the local flammability map induced by scene objects, letting us examine whether ProACT can coordinate global fire regulation with local semantic adaptation under more realistic fire dynamics.

Table~\ref{tab:3d_summary} and Figure~\ref{fig:dist3d} show that the 3D setting produces a more complex comparison than the 2D grid-world. Because propagation now depends on geometry and heterogeneous flammability, a fixed spread setting can perform well under some capability levels but fail under others. \emph{StaticEasy} achieves relatively high OT at $\epsilon=0.20$ and $0.40$ ($40.0\%$ and $65.0\%$) but drops to $23.0\%$ at $\epsilon=0.60$ as completion times shift toward the horizon.
\emph{StaticHard} shows the reverse pattern: competitive at $\epsilon=0.05$ ($41.0\%$) but dropping to $1.0\%$ at $\epsilon=0.60$. \emph{Reactive} peaks in the mid-range ($66.0\%$ at $\epsilon=0.40$) but is unstable under the most inconsistent behavior, falling to $23.0\%$ at $\epsilon=0.60$. By contrast, \emph{ProACT} is strongest at the extreme capability levels, achieving the highest OT at $\epsilon=0.05$, $\epsilon=0.20$ and $\epsilon=0.60$ ($49.0\%$, $48.0\%$, and $70.0\%$). The improvement at $\epsilon=0.60$ is particularly important: ProACT reduces the mean completion time to $98.10$ steps, versus $134.53$ (StaticEasy) and $133.97$ (Reactive). 
In this richer setting, ProACT operates across both global and local scales and remains robust across capability levels, particularly when simulated trainee behavior is inconsistent.
More broadly, these results highlight that difficulty is relative to trainee behavior: as behavior becomes less consistent, even StaticEasy can become over-challenging, while StaticHard amplifies overload; ProACT provides more stable regulation by tracking both short-term pacing and the target fire-load trajectory.

Fig.~\ref{fig:3d_completion_nums} illustrates how these differences unfold over time. In the 3D setting, heterogeneous object flammability causes active fire load to evolve unevenly across regions, making fixed spread settings less reliable. \textit{StaticEasy} can suppress early fire growth but often allows fires to persist beyond the target window under inconsistent behavior, while \textit{StaticHard} tends to amplify local growth and overload the scenario. \textit{Reactive} mitigates some deviations through global updates, but its response can lag behind localized fire dynamics. In contrast, \textit{ProACT} coordinates global fire-spread regulation with object-aware flammability updates, enabling it to control both overall pacing and local ignition likelihood. This is most evident at $\epsilon=0.60$, where ProACT completes near the target time while other methods remain outside the target window or terminate late. These traces suggest that ProACT maintains appropriate training pace by regulating fire dynamics across both temporal and spatial scales.

\section{VR User Study}
We conducted an exploratory VR user study to examine how participants experienced ProACT in an interactive VR firefighter training scenario, focusing on whether performance-driven adaptation could maintain an appropriate level of challenge while preserving engagement, perceived realism, and trust in the adaptive behavior.

\subsection{Study Design}
We used a within-subjects design in which each participant experienced all four system conditions---StaticEasy, StaticHard, Reactive, and ProACT---with condition order randomized per participant to reduce ordering effects. Each trial used the same task objective: extinguish the fire in the virtual environment. Following the regulation goal from our simulation study, the target completion window was set to $60$--$120$\,s, and a trial ended when the participant extinguished all active fires or reached a $150$\,s time limit. We recorded task completion time and the number of active fire cells extinguished during each trial. After each condition, participants completed a short post-condition questionnaire of eight 7-point Likert items (Table~\ref{tab:userstudy_questionnaire}) measuring engagement, perceived realism, perceived adaptation, trust in adaptation, challenge appropriateness, mental-effort appropriateness, perceived training relevance, and willingness to reuse the system. Each session lasted about $30$ minutes, including setup, familiarization, formal trials, and questionnaires.

\begin{table}[t]
\centering
\caption{Post-condition questionnaire items used in the user study.}
\label{tab:userstudy_questionnaire}
\footnotesize
\setlength{\tabcolsep}{3pt}
\renewcommand{\arraystretch}{1.05}
\begin{tabularx}{\columnwidth}{@{} llX @{}}
\toprule
\textbf{Item} & \textbf{Construct} & \textbf{Statement} \\
\midrule
Q1 & Engagement & I found the training scenario engaging and interesting. \\
Q2 & Realism & The virtual fire environment felt realistic enough for me to respond as if the situation mattered. \\
Q3 & Adaptation & The system appeared to adjust the fire behavior in response to my actions. \\
Q4 & Trust & I trusted that the system adapted the difficulty in a reasonable way based on my performance. \\
Q5 & Challenge & The difficulty level in this condition was appropriate for effective learning. \\
Q6 & Effort & The mental effort required during this condition was appropriate---neither too high nor too low. \\
Q7 & Relevance & This training condition helped me develop skills relevant to real firefighting situations. \\
Q8 & Reuse & I would choose to use this type of XR training again in the future. \\
\bottomrule
\end{tabularx}

\vspace{2pt}
\raggedright \scriptsize \textit{Note: All items were rated on a 7-point Likert scale from 1 = Strongly Disagree to 7 = Strongly Agree.}
\end{table}

\subsection{Participants and Ethics}
We recruited eight participants from the university community for this evaluation (aged 23--30 years, mean age $26.4$). Before the study, each participant completed a pre-study questionnaire covering demographic information and prior experience with XR technologies and immersive 3D environments, including VR applications, MR headsets, and 3D first-person video games. Participants self-reported varying levels of XR experience: two beginners, two intermediate users, and four experts. The study protocol was approved by our institution's Institutional Review Board (IRB). Each participant received a gift card as compensation.

\subsection{Procedure}
At the start of the study, participants received a brief introduction to the project, the training task, and the purpose of the experiment, and then completed the pre-study questionnaire. Next, they completed a familiarization session in the VR environment. This session did not include adaptive difficulty regulation and was used only to help participants learn the basic interactions---navigation, fire observation, and extinguisher-based suppression. Once participants indicated they were comfortable with the controls, they began the formal trials, completing all four conditions in randomized order. During each trial, participants extinguished all active fires using the virtual extinguisher while the assigned condition determined how the fire evolved; a trial ended when all fires were extinguished or the $150$\,s limit was reached. Between trials, participants rested for at least one minute (longer if needed). After each condition they completed the post-condition questionnaire, and after all conditions they finished the session.

\subsection{Results}

\begin{figure}[t]
    \centering
    \includegraphics[width=\linewidth]{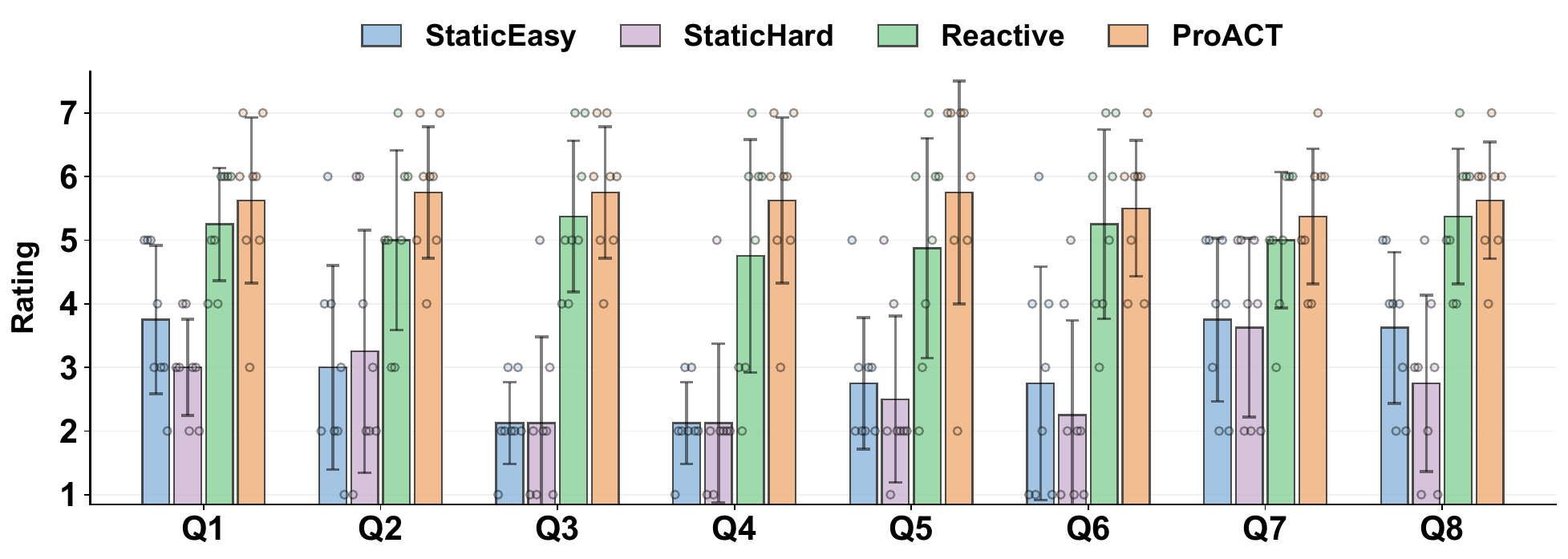}
    \Description{Ratings on eight questionnaire items for four study conditions, showing means, standard deviations, and individual participant responses.}
    \caption{Post-condition questionnaire ratings across the four conditions (Q1--Q8: engagement, realism, adaptation, trust, challenge, effort, relevance, reuse). Bars show means, error bars standard deviations, and dots individual participant ratings (7-point Likert scale).}
    \label{fig:user_study}
    \vspace{-5mm}
\end{figure}

\subsubsection{Questionnaire Results}
As shown in Figure~\ref{fig:user_study}, ProACT received positive ratings across all dimensions, with mean scores above $5.3/7$ for every item, and achieved the
highest mean rating among the four conditions on every item---engagement (Q1), realism (Q2), adaptation (Q3), trust (Q4), challenge (Q5), effort (Q6), relevance (Q7), and reuse intention (Q8). This suggests participants perceived ProACT as more engaging, more adaptive, and better able to maintain an appropriate level of challenge than the baselines. The largest differences were over the two static baselines, which scored lower across most dimensions---especially perceived adaptation (Q3), trust (Q4), challenge (Q5), and effort (Q6)---consistent with the design of fixed-difficulty scenarios: because fire behavior does not respond to performance, the same scenario can become too easy or too hard depending on the participant. \emph{Reactive} also received relatively positive ratings, particularly on perceived adaptation (Q3), as expected given its dynamic fire response; however, ProACT received higher mean ratings than Reactive on all items, with the largest gaps on trust (Q4), challenge appropriateness (Q5), and reuse intention (Q8). This indicates participants valued not merely that the fire behavior changed, but that the resulting difficulty felt reasonable and useful for training. Overall, these results support our design goal of maintaining a ``just-right'' level of difficulty through performance-driven environment regulation, while suggesting that adaptive fire behavior can improve engagement and perceived training relevance in VR firefighter training.

\begin{figure*}[t]
    \centering
    \includegraphics[width=.8\linewidth]{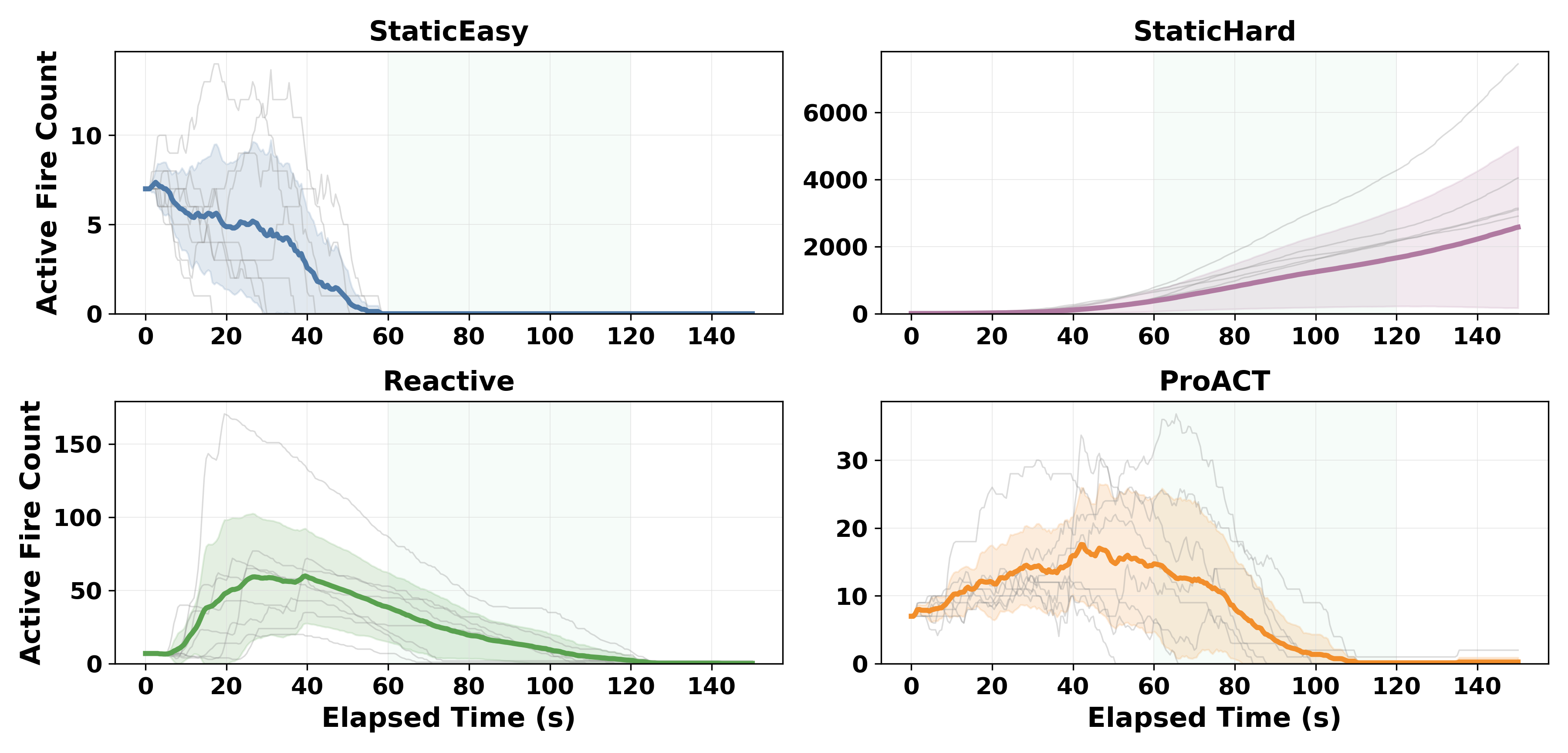}
    \Description{Active fire-load trajectories for the four user-study conditions, showing individual trials, mean curves, standard deviations, and the target completion window.}
    \caption{Active fire-load progression over time in the user-study trials, one panel per condition. Thin gray lines are individual trials, colored lines the mean active fire count, and shaded bands $\pm1$ standard deviation. The vertical shaded region marks the $60$--$120$\,s target window. ProACT generally drives the active fire load to zero within the target completion window; StaticEasy ends too early and StaticHard grows out of control.}
    \label{fig:user_time}
\end{figure*}

\subsubsection{Log-Based Fire-Load Analysis}
The logged fire-load trajectories (Figure~\ref{fig:user_time}) corroborate the ratings. The fixed baselines show two opposite failure patterns. In \emph{StaticEasy}, the mean active fire count quickly decreases to zero before the target window, with most trials ending around $60$\,s or earlier, indicating that the easy setting often made the task under-challenging and left little opportunity for sustained practice. In contrast, \emph{StaticHard} caused the fire load to grow continuously: the mean active fire count exceeded $2{,}000$ by the end of a session, and some trials grew much higher, suggesting the scenario became difficult to control and often exceeded the $60$--$120$\,s window. \emph{Reactive} showed better control than the static baselines---the mean fire load rose early, peaked around $50$--$60$ active fires, and gradually decreased toward zero near the end of the window---but its trajectory was smooth and slow-changing, suggesting only coarse threshold-based adjustment that did not closely track moment-to-moment changes in performance. \emph{ProACT} produced a distinct pattern: the fire load rose moderately during the early phase, fluctuated as the system adapted to ongoing suppression progress, and then decreased to zero within the target window. Compared with Reactive, ProACT's mean curve shows more local adjustment rather than a single gradual decline, which is consistent with its performance-driven regulation. Overall, the trajectories support the questionnaire findings: StaticEasy tended to finish too early, StaticHard often became uncontrollable, Reactive provided partial but coarse adaptation, and ProACT better maintained the scenario within the intended range by continuously adjusting fire behavior during the task.

\section{Discussion}

\subsection{Implications}
Our results highlight a practical approach to adapting dynamic immersive training without relying on predefined difficulty levels or explicit user-skill models. Rather than switching tasks, changing interaction rules, or modifying only instructional feedback, ProACT regulates the simulated fire dynamics themselves, allowing challenge to evolve continuously within the same training scenario as trainee performance changes. Regulation is driven by observable task progress, including suppression pacing and remaining fire load, so the system can respond directly to whether progression is ahead of or behind the target without first inferring latent skill or intent. ProACT further separates continuous control from semantic grounding: the real-time global controller regulates overall fire progression, while periodic semantic reasoning identifies where bounded local adjustments are appropriate from scene context. Importantly, the LLM is therefore not responsible for continuously controlling difficulty; its role is restricted to contextualizing local adaptation, while performance-driven signals govern the direction of regulation. This division preserves the responsiveness of numerical feedback control while incorporating scene semantics where spatial decisions are required. More broadly, ProACT shows how an adaptive XR system can keep the task objective and interaction loop unchanged while modifying the underlying process that generates challenge, making the simulated environment itself part of the feedback loop between trainee performance and scenario progression.

\subsection{Limitations}
This work has three main limitations. First, ProACT currently targets individual trainees, whereas real-world firefighting is inherently collaborative. Extending the framework to multi-firefighter scenarios would require regulating challenge at the team level, accounting for role differences and coordination, and synchronizing adaptive events across multiple HMD clients in a shared virtual environment; it remains open whether difficulty should be calibrated to the team, to individual trainees, or to roles within the team. Second, our simulation-based evaluation relies on simplified trainee behavior models and controlled fire dynamics. Although the $\epsilon$-greedy simulated trainee enables repeatable capability levels, it does not capture the full range of human decision-making, stress responses, or navigation strategies in real emergencies. Completion time also serves as an operational measure of task pacing rather than a comprehensive measure of perceived challenge or learning effectiveness. Third, our exploratory VR study is modest in scale ($n=8$), draws participants from a university community, and uses custom Likert items rather than fully standardized instruments. Future studies involving broader participant populations and validated measures of workload, presence, and cybersickness, together with objective learning-transfer outcomes, could provide a more comprehensive assessment of performance-driven environment regulation.

\subsection{Broader Applicability}
ProACT was developed for VR firefighter training. Its performance-driven environment regulation may apply more broadly to immersive training tasks in which difficulty emerges from dynamic scenario behavior. The framework is most applicable when task progression can be observed online, environment dynamics expose controllable parameters, and local adaptations can be constrained using scene context. Such conditions arise in domains including emergency response, disaster evacuation, and industrial safety, where evolving hazards or task pressure contribute directly to training difficulty. Rather than treating adaptation primarily as a change in interface feedback or task selection, ProACT suggests a complementary design direction in which the simulated environment itself becomes part of the adaptive training loop.

\section{Conclusion}
We presented \textbf{ProACT}, a performance-driven adaptive training system for VR firefighter training. Prior work has provided safe and immersive practice opportunities, but many systems rely on fixed scenarios or limited adaptation that may not sustain an appropriate level of challenge for trainees of different skill levels. To address this, ProACT continuously regulates scenario difficulty through closed-loop interactions between trainee performance and simulated fire behavior, coupling a real-time global fire-dynamics controller with a periodic, LLM-based local flammability-reasoning module. Across controlled 2D and 3D simulations and a VR user study, ProACT better maintained a ``just-right'' challenge level and improved perceived challenge appropriateness, trust, and reuse intention compared with adaptive and non-adaptive baselines. Our work highlights the potential of performance-driven adaptive environments for immersive training and may inspire future systems in domains such as emergency response, industrial safety, and procedural skill rehearsal.

\bibliographystyle{ACM-Reference-Format}
\bibliography{main}

@article{scorgie2024virtual,
  title={Virtual reality for safety training: A systematic literature review and meta-analysis},
  author={Scorgie, D and Feng, Z and Paes, D and Parisi, F and Yiu, TW and Lovreglio, R},
  journal={Safety science},
  volume={171},
  pages={106372},
  year={2024},
  publisher={Elsevier}
}

@article{killeen2024stand,
  title={Stand in surgeon’s shoes: virtual reality cross-training to enhance teamwork in surgery},
  author={Killeen, Benjamin D and Zhang, Han and Wang, Liam J and Liu, Zixuan and Kleinbeck, Constantin and Rosen, Michael and Taylor, Russell H and Osgood, Greg and Unberath, Mathias},
  journal={International journal of computer assisted radiology and surgery},
  volume={19},
  number={6},
  pages={1213--1222},
  year={2024},
  publisher={Springer}
}

@inproceedings{chen2025bridging,
  title={Bridging Simulation and Reality: Augmented Virtuality for Mass Casualty Triage Training-From Landscape Analysis to Empirical Insights},
  author={Chen, Yang and Fennedy, Katherine and Zhang, Jiayi and Sim, Yong Jie and Zheng, Clement and Yen, Ching Chiuan},
  booktitle={Proceedings of the 2025 CHI Conference on Human Factors in Computing Systems},
  pages={1--28},
  year={2025}
}

@misc{bls_firefighters,
  author       = {{Bureau of Labor Statistics, U.S. Department of Labor}},
  title        = {Occupational Outlook Handbook: Firefighters},
  howpublished = {\url{https://www.bls.gov/ooh/protective-service/firefighters.htm}},
  year         = {2026}, 
  note         = {Accessed: 2026-03-11}
}

@techreport{nfpa_injury_report,
  author      = {Richard Campbell and Shelby Hall and Jay Petrillo},
  title       = {Firefighter Injuries in the United States},
  institution = {National Fire Protection Association (NFPA)},
  year        = {2026},
  type        = {Research Report}
}

@article{guadagnoli2004challenge,
  title={Challenge point: a framework for conceptualizing the effects of various practice conditions in motor learning},
  author={Guadagnoli, Mark A and Lee, Timothy D},
  journal={Journal of motor behavior},
  volume={36},
  number={2},
  pages={212--224},
  year={2004},
  publisher={Taylor \& Francis}
}

@article{bonaiuto2016optimal,
  title={Optimal experience and personal growth: Flow and the consolidation of place identity},
  author={Bonaiuto, Marino and Mao, Yanhui and Roberts, Scott and Psalti, Anastasia and Ariccio, Silvia and Ganucci Cancellieri, Uberta and Csikszentmihalyi, Mihaly},
  journal={Frontiers in psychology},
  volume={7},
  pages={1654},
  year={2016},
  publisher={Frontiers Media SA}
}

@article{mehta2025effectiveness,
  title={Effectiveness of training under stress in immersive VR: an investigation of firefighter performance, gaze entropy, and pupillometry},
  author={Mehta, Ranjana K and Kang, John and Shi, Yangming and Du, Jing},
  journal={Frontiers in Virtual Reality},
  volume={6},
  pages={1542507},
  year={2025},
  publisher={Frontiers Media SA}
}

@article{engelbrecht2019swot,
  title={A SWOT analysis of the field of virtual reality for firefighter training},
  author={Engelbrecht, Hendrik and Lindeman, Robert W and Hoermann, Simon},
  journal={Frontiers in Robotics and AI},
  volume={6},
  pages={101},
  year={2019},
  publisher={Frontiers Media SA}
}

@article{wheeler2021human,
  title={Human factors research in immersive virtual reality firefighter training: A systematic review},
  author={Wheeler, Steven G and Engelbrecht, Hendrik and Hoermann, Simon},
  journal={Frontiers in Virtual Reality},
  volume={2},
  pages={671664},
  year={2021},
  publisher={Frontiers Media SA}
}

@inproceedings{clifford2019creating,
  title={Creating a stressful decision making environment for aerial firefighter training in virtual reality},
  author={Clifford, Rory MS and Jung, Sungchul and Hoermann, Simon and Billinghurst, Mark and Lindeman, Robert W},
  booktitle={2019 IEEE Conference on virtual reality and 3d user interfaces (VR)},
  pages={181--189},
  year={2019},
  organization={IEEE}
}

@article{verniani2024features,
  title={Features of adaptive training algorithms for improved complex skill acquisition},
  author={Verniani, Alessandro and Galvin, Ellery and Tredinnick, Sandra and Putman, Esther and Vance, Eric A and Clark, Torin K and Anderson, Allison P},
  journal={Frontiers in Virtual Reality},
  volume={5},
  pages={1322656},
  year={2024},
  publisher={Frontiers Media SA}
}

@article{zahabi2020adaptive,
  title={Adaptive virtual reality-based training: a systematic literature review and framework},
  author={Zahabi, Maryam and Abdul Razak, Ashiq Mohammed},
  journal={Virtual Reality},
  volume={24},
  number={4},
  pages={725--752},
  year={2020},
  publisher={Springer}
}

@article{butaslac2022systematic,
  title={Systematic review of augmented reality training systems},
  author={Butaslac, Isidro III Mendoza and Fujimoto, Yuichiro and Sawabe, Taishi and Kanbara, Masayuki and Kato, Hirokazu},
  journal={IEEE Transactions on Visualization and Computer Graphics},
  volume={29},
  number={12},
  pages={5062--5082},
  year={2022},
  publisher={IEEE}
}

@article{vaughan2016overview,
  title={An overview of self-adaptive technologies within virtual reality training},
  author={Vaughan, Neil and Gabrys, Bodgan and Dubey, Venketesh N},
  journal={Computer Science Review},
  volume={22},
  pages={65--87},
  year={2016},
  publisher={Elsevier}
}

@article{oliveira2024exploring,
  title={Exploring immersive multimodal virtual reality training, affective states, and ecological validity in healthy firefighters: quasi-experimental study},
  author={Oliveira, Joana and Dias, Joana Aires and Correia, Rita and Pinheiro, Raquel and Reis, V{\'\i}tor and Sousa, Daniela and Agostinho, Daniel and Sim{\~o}es, Marco and Castelo-Branco, Miguel},
  journal={JMIR Serious Games},
  volume={12},
  number={1},
  pages={e53683},
  year={2024},
  publisher={JMIR Publications Inc., Toronto, Canada}
}

@article{cha2012virtual,
  title={A virtual reality based fire training simulator integrated with fire dynamics data},
  author={Cha, Moohyun and Han, Soonhung and Lee, Jaikyung and Choi, Byungil},
  journal={Fire safety journal},
  volume={50},
  pages={12--24},
  year={2012},
  publisher={Elsevier}
}

@inproceedings{backlund2007sidh,
  title={Sidh--a game based firefighter training simulation},
  author={Backlund, Per and Engstrom, Henrik and Wijkmark, Cecilia Hammar and Johannesson, Mikael and Lebram, Mikael},
  booktitle={2007 11th International Conference Information Visualization (IV'07)},
  pages={899--907},
  year={2007},
  organization={IEEE}
}

@article{wang2024mrehab,
  title={Mrehab: A mixed reality rehabilitation system supporting integrated speech and hand training},
  author={Wang, Qi and Luo, Can and Yang, Shuai and Xu, Shuo and Yang, Yibo and Jia, Jie and Yu, Bin},
  journal={Proceedings of the ACM on Interactive, Mobile, Wearable and Ubiquitous Technologies},
  volume={8},
  number={3},
  pages={1--23},
  year={2024},
  publisher={ACM New York, NY, USA}
}

@inproceedings{pietschmann2024use,
  title={Use of virtual reality for emergency service training},
  author={Pietschmann, Leon and Pickhardt, Fritz},
  booktitle={Extended Abstracts of the CHI Conference on Human Factors in Computing Systems},
  pages={1--7},
  year={2024}
}

@inproceedings{yu2024design,
  title={Design space of visual feedforward and corrective feedback in xr-based motion guidance systems},
  author={Yu, Xingyao and Lee, Benjamin and Sedlmair, Michael},
  booktitle={Proceedings of the 2024 CHI Conference on Human Factors in Computing Systems},
  pages={1--15},
  year={2024}
}

@inproceedings{cheng2024viscourt,
  title={Viscourt: In-situ guidance for interactive tactic training in mixed reality},
  author={Cheng, Liqi and Jia, Hanze and Yu, Lingyun and Wu, Yihong and Ye, Shuainan and Deng, Dazhen and Zhang, Hui and Xie, Xiao and Wu, Yingcai},
  booktitle={Proceedings of the 37th Annual ACM Symposium on User Interface Software and Technology},
  pages={1--14},
  year={2024}
}

@inproceedings{studer2024open,
  title={An open-ended system in virtual reality for training machining skills},
  author={Studer, Kachina and Lie, Hing and Zhao, Zhen and Thomson, Ben and Turakhia, Dishita G and Liu, John},
  booktitle={Extended Abstracts of the CHI Conference on Human Factors in Computing Systems},
  pages={1--5},
  year={2024}
}

@article{lebrun2025mentor,
  title={Mentor-guided learning in immersive virtual environments: The impact of visual and haptic feedback on skill acquisition},
  author={Lebrun, Flavien and Simon, Cassandre and Boukezzi, Assia and Otmane, Samir and Chellali, Amine},
  journal={IEEE Transactions on Visualization and Computer Graphics},
  year={2025},
  publisher={IEEE}
}

@article{diller2022visual,
  title={Visual cue based corrective feedback for motor skill training in mixed reality: A survey},
  author={Diller, Florian and Scheuermann, Gerik and Wiebel, Alexander},
  journal={IEEE Transactions on Visualization and Computer Graphics},
  volume={30},
  number={7},
  pages={3121--3134},
  year={2022},
  publisher={IEEE}
}

@inproceedings{hunicke2005case,
  title={The case for dynamic difficulty adjustment in games},
  author={Hunicke, Robin},
  booktitle={Proceedings of the 2005 ACM SIGCHI International Conference on Advances in computer entertainment technology},
  pages={429--433},
  year={2005}
}

@article{zohaib2018dynamic,
  title={Dynamic difficulty adjustment (DDA) in computer games: A review},
  author={Zohaib, Mohammad},
  journal={Advances in Human-Computer Interaction},
  volume={2018},
  number={1},
  pages={5681652},
  year={2018},
  publisher={Wiley Online Library}
}

@article{aponte2011measuring,
  title={Measuring the level of difficulty in single player video games},
  author={Aponte, Maria-Virginia and Levieux, Guillaume and Natkin, Stephane},
  journal={Entertainment Computing},
  volume={2},
  number={4},
  pages={205--213},
  year={2011},
  publisher={Elsevier}
}

@inproceedings{missura2009player,
  title={Player modeling for intelligent difficulty adjustment},
  author={Missura, Olana and G{\"a}rtner, Thomas},
  booktitle={International Conference on Discovery Science},
  pages={197--211},
  year={2009},
  organization={Springer}
}

@inproceedings{pfau2020enemy,
  title={Enemy within: Long-term motivation effects of deep player behavior models for dynamic difficulty adjustment},
  author={Pfau, Johannes and Smeddinck, Jan David and Malaka, Rainer},
  booktitle={Proceedings of the 2020 CHI Conference on Human Factors in Computing Systems},
  pages={1--10},
  year={2020}
}

@article{pfau2022dungeons,
  title={Dungeons \& replicants ii: automated game balancing across multiple difficulty dimensions via deep player behavior modeling},
  author={Pfau, Johannes and Liapis, Antonios and Yannakakis, Georgios N and Malaka, Rainer},
  journal={IEEE Transactions on Games},
  volume={15},
  number={2},
  pages={217--227},
  year={2022},
  publisher={IEEE}
}

@inproceedings{zook2012skill,
  title={Skill-based mission generation: A data-driven temporal player modeling approach},
  author={Zook, Alexander and Lee-Urban, Stephen and Drinkwater, Michael R and Riedl, Mark O},
  booktitle={Proceedings of the The third workshop on Procedural Content Generation in Games},
  pages={1--8},
  year={2012}
}

@inproceedings{biemer2023dynamic,
  title={Dynamic difficulty adjustment via procedural level generation guided by a Markov decision process for platformers and roguelikes},
  author={Biemer, Colan F},
  booktitle={Proceedings of the AAAI Conference on Artificial Intelligence and Interactive Digital Entertainment},
  volume={19},
  number={1},
  pages={436--439},
  year={2023}
}

@article{moschovitis2022keep,
  title={Keep calm and aim for the head: Biofeedback-controlled dynamic difficulty adjustment in a horror game},
  author={Moschovitis, Paraschos and Denisova, Alena},
  journal={IEEE Transactions on Games},
  volume={15},
  number={3},
  pages={368--377},
  year={2022},
  publisher={IEEE}
}

@inproceedings{correia2024adaptive,
  title={Adaptive virtual reality solutions: a literature review},
  author={Correia, Pedro Henrique Barcha},
  booktitle={Proceedings of the 26th Symposium on Virtual and Augmented Reality},
  pages={1--10},
  year={2024}
}

@article{aguilar2023design,
  title={Design and evaluation of an adaptive virtual reality training system},
  author={Aguilar Reyes, C{\'e}sar Iv{\'a}n and Wozniak, David and Ham, Angel and Zahabi, Maryam},
  journal={Virtual Reality},
  volume={27},
  number={3},
  pages={2509--2528},
  year={2023},
  publisher={Springer}
}

@article{liao2025focus,
  title={Focus-driven augmented feedback: Enhancing focus and maintaining engagement in upper limb virtual reality rehabilitation},
  author={Liao, Kai-Lun and Huang, Mengjie and Shi, Jiajia and Chen, Min and Yang, Rui},
  journal={IEEE Transactions on Visualization and Computer Graphics},
  year={2025},
  publisher={IEEE}
}

@article{chu2025design,
  title={Design and evaluation of AR-based adaptive human-computer interaction cognitive training},
  author={Chu, Man and Qu, Jing and Zou, Tan and Li, Qinbiao and Bu, Lingguo and Shen, Yiran},
  journal={International Journal of Human-Computer Studies},
  volume={199},
  pages={103504},
  year={2025},
  publisher={Elsevier}
}

@article{chopra2025engagements,
  title={Engagements with Generative AI and Personal Health Informatics: Opportunities for Planning, Tracking, Reflecting, and Acting around Personal Health Data},
  author={Chopra, Shaan and Juarez, Katherine and Fogarty, James and Munson, Sean A},
  journal={Proceedings of the ACM on Interactive, Mobile, Wearable and Ubiquitous Technologies},
  volume={9},
  number={3},
  pages={1--33},
  year={2025},
  publisher={ACM New York, NY, USA}
}

@article{li2025vital,
  title={Vital Insight: Assisting Experts' Context-Driven Sensemaking of Multi-modal Personal Tracking Data Using Visualization and Human-in-the-Loop LLM},
  author={Li, Jiachen and Li, Xiwen and Steinberg, Justin and Choube, Akshat and Yao, Bingsheng and Xu, Xuhai and Wang, Dakuo and Mynatt, Elizabeth and Mishra, Varun},
  journal={Proceedings of the ACM on Interactive, Mobile, Wearable and Ubiquitous Technologies},
  volume={9},
  number={3},
  pages={1--37},
  year={2025},
  publisher={ACM New York, NY, USA}
}

@article{jiang2024haigen,
  title={HAIGEN: towards human-AI collaboration for facilitating creativity and style generation in fashion design},
  author={Jiang, Jianan and Wu, Di and Deng, Hanhui and Long, Yidan and Tang, Wenyi and Li, Xiang and Liu, Can and Jin, Zhanpeng and Zhang, Wenlei and Qi, Tangquan},
  journal={Proceedings of the ACM on Interactive, Mobile, Wearable and Ubiquitous Technologies},
  volume={8},
  number={3},
  pages={1--27},
  year={2024},
  publisher={ACM New York, NY, USA}
}

@article{nepal2024mindscape,
  title={MindScape study: integrating LLM and behavioral sensing for personalized AI-driven journaling experiences},
  author={Nepal, Subigya and Pillai, Arvind and Campbell, William and Massachi, Talie and Heinz, Michael V and Kunwar, Ashmita and Choi, Eunsol Soul and Xu, Xuhai and Kuc, Joanna and Huckins, Jeremy F and others},
  journal={Proceedings of the ACM on interactive, mobile, wearable and ubiquitous technologies},
  volume={8},
  number={4},
  pages={1--44},
  year={2024},
  publisher={ACM New York, NY, USA}
}

@inproceedings{chen2025origami,
  title={Origami sensei: A mixed reality ai-assistant},
  author={Chen, Qiyu and Mishra, Richa and Purnamasari, Lia Sparingga and EL-Zanfaly, Dina and Kitani, Kris},
  booktitle={Proceedings of the 2025 CHI conference on human factors in computing systems},
  pages={1--18},
  year={2025}
}

@inproceedings{shi2025caring,
  title={Caring-ai: Towards authoring context-aware augmented reality instruction through generative artificial intelligence},
  author={Shi, Jingyu and Jain, Rahul and Chi, Seunggeun and Doh, Hyungjun and Chi, Hyung-gun and Quinn, Alexander J and Ramani, Karthik},
  booktitle={Proceedings of the 2025 CHI conference on human factors in computing systems},
  pages={1--23},
  year={2025}
}

@inproceedings{wan2024building,
  title={Building llm-based ai agents in social virtual reality},
  author={Wan, Hongyu and Zhang, Jinda and Suria, Abdulaziz Arif and Yao, Bingsheng and Wang, Dakuo and Coady, Yvonne and Prpa, Mirjana},
  booktitle={Extended Abstracts of the CHI Conference on Human Factors in Computing Systems},
  pages={1--7},
  year={2024}
}

@inproceedings{jiang2026partinteractor,
  title={PartInteractor: Intent-Driven Part-Aware 3D Authoring for Continuous Co-Creation in XR},
  author={Jiang, Jianan and Li, Bin},
  booktitle={Proceedings of the 39th Annual ACM Symposium on User Interface Software and Technology},
  pages={1--13},
  year={2026}
}

@inproceedings{cho2024minexr,
  title={Minexr: mining personalized extended reality interfaces},
  author={Cho, Hyunsung and Yan, Yukang and Todi, Kashyap and Parent, Mark and Smith, Missie and Jonker, Tanya R and Benko, Hrvoje and Lindlbauer, David},
  booktitle={Proceedings of the 2024 CHI Conference on Human Factors in Computing Systems},
  pages={1--17},
  year={2024}
}

@inproceedings{lee2024gazepointar,
  title={GazePointAR: A context-aware multimodal voice assistant for pronoun disambiguation in wearable augmented reality},
  author={Lee, Jaewook and Wang, Jun and Brown, Elizabeth and Chu, Liam and Rodriguez, Sebastian S and Froehlich, Jon E},
  booktitle={Proceedings of the 2024 CHI Conference on Human Factors in Computing Systems},
  pages={1--20},
  year={2024}
}

@inproceedings{li2024situationadapt,
  title={Situationadapt: Contextual ui optimization in mixed reality with situation awareness via llm reasoning},
  author={Li, Zhipeng and Gebhardt, Christoph and Inglin, Yves and Steck, Nicolas and Streli, Paul and Holz, Christian},
  booktitle={Proceedings of the 37th Annual ACM Symposium on User Interface Software and Technology},
  pages={1--13},
  year={2024}
}

@article{singh2025openai,
  title={Openai gpt-5 system card},
  author={Singh, Aaditya and Fry, Adam and Perelman, Adam and Tart, Adam and Ganesh, Adi and El-Kishky, Ahmed and McLaughlin, Aidan and Low, Aiden and Ostrow, AJ and Ananthram, Akhila and others},
  journal={arXiv preprint arXiv:2601.03267},
  year={2025}
}

@inproceedings{sepulveda2019exploring,
  title={Exploring dynamic difficulty adjustment in videogames},
  author={Sepulveda, Gabriel K and Besoain, Felipe and Barriga, Nicolas A},
  booktitle={2019 IEEE CHILEAN Conference on Electrical, Electronics Engineering, Information and Communication Technologies (CHILECON)},
  pages={1--6},
  year={2019},
  organization={IEEE}
}

@article{lin2025vrshuttlecock,
  title={VR-Shuttlecock: Toward Enhancing Older Adults' Balance Through Kicking Shuttlecock in VR with Multi-Sensory Feedback},
  author={Lin, Fan and Liang, Yiqi and Wei, Qianjie and Li, Mingshuo and Jiang, Chutian and Fan, Mingming},
  journal={Proceedings of the ACM on Interactive, Mobile, Wearable and Ubiquitous Technologies},
  volume={9},
  number={4},
  pages={1--27},
  year={2025},
  publisher={ACM New York, NY, USA}
}

\end{document}